\documentclass[10pt,journal,compsoc]{IEEEtran}

\usepackage{color}
\usepackage[normalem]{ulem}

\ifCLASSOPTIONcompsoc
  \usepackage[nocompress]{cite}
\else
  \usepackage{cite}
\fi

\ifCLASSINFOpdf
   \usepackage[pdftex]{graphicx}
   \graphicspath{{../pdf/}{../jpeg/}{./figs/}{./figures}}
\else
\fi

\usepackage{graphicx}
\ifCLASSINFOpdf

\else

\fi

\usepackage[cmex10]{amsmath}
\usepackage{array}
\usepackage{multirow}
\usepackage{lineno}
\usepackage{url}
\usepackage{amsmath}

\def\0{{\bf 0}}
\def\1{{\bf 1}}

\def\etal{{\em et al.}}
\def\eg{{\em e.g.}}
\def\ie{{\em i.e.}}

\def\etal{{\em et al.\/}\,}

\usepackage{amsmath}
\usepackage{amssymb}
\usepackage{multirow}
\usepackage{array}
\usepackage{booktabs}
\usepackage{multirow}
\usepackage{bm}
\usepackage{graphicx}
\usepackage{epstopdf}
\usepackage{subfigure}
\usepackage{makecell}
\usepackage[pagebackref=false,breaklinks=true,letterpaper=true,colorlinks,bookmarks=false]{hyperref}
\usepackage{cite}
\usepackage{multirow}
\usepackage{color}
\usepackage{CJK}
\usepackage{soul}
\usepackage[numbers,sort&compress]{natbib}
\usepackage{hyperref}

\usepackage[colorinlistoftodos]{todonotes}
\def\etal{{\em et al.}}
\def\eg{{\em e.g.}}
\def\ie{{\em i.e.}}

\begin{document}

\title{Hyperbolic Kernel Graph Neural Networks for Neurocognitive Decline Analysis from Multimodal Brain Imaging}

\author{Meimei~Yang, 
Yongheng Sun, 
Qianqian~Wang,
Andrea~Bozoki, 
Maureen Kohi, 
Mingxia~Liu,~\IEEEmembership{Senior Member,~IEEE}

\IEEEcompsocitemizethanks{
\IEEEcompsocthanksitem M.~Yang, Y.~Sun, Q.~Wang, M. Kohi, and M.~Liu are with the Department of Radiology and Biomedical Research Imaging Center (BRIC), University of North Carolina at Chapel Hill, Chapel Hill, NC 27599, USA. 
A.~Bozoki is with the Department of Neurology, University of North Carolina at Chapel Hill, Chapel Hill, NC 27599, USA. 
\IEEEcompsocthanksitem 
Corresponding author: M.~Liu (Email: mingxia\_liu@med.unc.edu). 
\protect\\
}
}
{}

\IEEEtitleabstractindextext{%
\begin{abstract}
Multimodal neuroimages, such as diffusion tensor imaging (DTI) and resting-state functional MRI (fMRI), 
offer complementary perspectives on brain activities by capturing structural or functional interactions among brain regions. 
While existing studies suggest that fusing these multimodal data helps detect abnormal brain activity caused by neurocognitive decline, they are generally implemented in Euclidean space and can't effectively capture intrinsic hierarchical organization of structural/functional 
brain networks.  
This paper presents a hyperbolic kernel graph fusion (HKGF) framework for 
neurocognitive decline analysis with multimodal neuroimages. 
It consists of a \emph{multimodal graph construction module}, a \emph{graph representation learning module} that encodes brain graphs in hyperbolic space  
through a family of hyperbolic kernel graph neural networks (HKGNNs), a \emph{cross-modality coupling module} that enables effective multimodal data fusion, 
and 
a \emph{hyperbolic neural network} for downstream predictions. 
Notably, HKGNNs represent graphs in hyperbolic space to capture both local and global dependencies among brain regions while preserving the hierarchical structure of brain networks. 
Extensive experiments involving over 4,000 subjects with DTI and/or fMRI data suggest the superiority of HKGF over state-of-the-art methods in two neurocognitive decline prediction tasks. 
HKGF is a general framework for multimodal data analysis, facilitating objective quantification of structural/functional brain connectivity changes associated with neurocognitive decline.

\end{abstract}

}

\maketitle

\IEEEdisplaynontitleabstractindextext

\IEEEpeerreviewmaketitle

\IEEEraisesectionheading{\section{Introduction}\label{S1}}


\IEEEPARstart{M}{ultimodal} neuroimaging data, such as diffusion tensor imaging (DTI), resting-state functional MRI (fMRI) and arterial spin labeling (ASL) provide complementary views of the brain~\cite{mulkern2006complementary}. 
While DTI characterizes the structural connectivity (SC) between brain regions-of-interest (ROIs)~\cite{basser1994mr}, 
fMRI and ASL capture the functional connectivity (FC) among ROIs by measuring correlated blood-oxygen-level-dependent (BOLD) signal fluctuations and regional cerebral blood flow, respectively~\cite{biswal1995functional,fox2007spontaneous,detre1999perfusion,alsop2015recommended}. 
\if false
Recent studies have demonstrated that integrating these multimodal neuroimaging data 
helps improve the identification and interpretation of abnormal brain connectivity patterns in clinical populations~\cite{zhu2014fusing}. 
In particular, multimodal neuroimaging fusion has proven valuable in neurocognitive decline analysis, where subtle alterations in structural, functional, or perfusion brain networks contribute to cognitive decline~\cite{zhou2010divergent}.
\fi 
Recent studies have shown that integrating these multimodal neuroimaging data improves the detection of abnormal brain connectivity patterns associated with neurocognitive decline~\cite{zhu2014fusing}, where subtle changes in structural, functional, or perfusion brain networks contribute to cognitive decline~\cite{zhou2010divergent}. 

\begin{figure*}[!t]
\setlength{\abovecaptionskip}{0pt}
\setlength{\belowcaptionskip}{0pt}
\setlength{\abovedisplayskip}{0pt}
\setlength\belowdisplayskip{0pt}
\setlength{\abovecaptionskip}{0pt}
\centering
\includegraphics[width=1\textwidth]{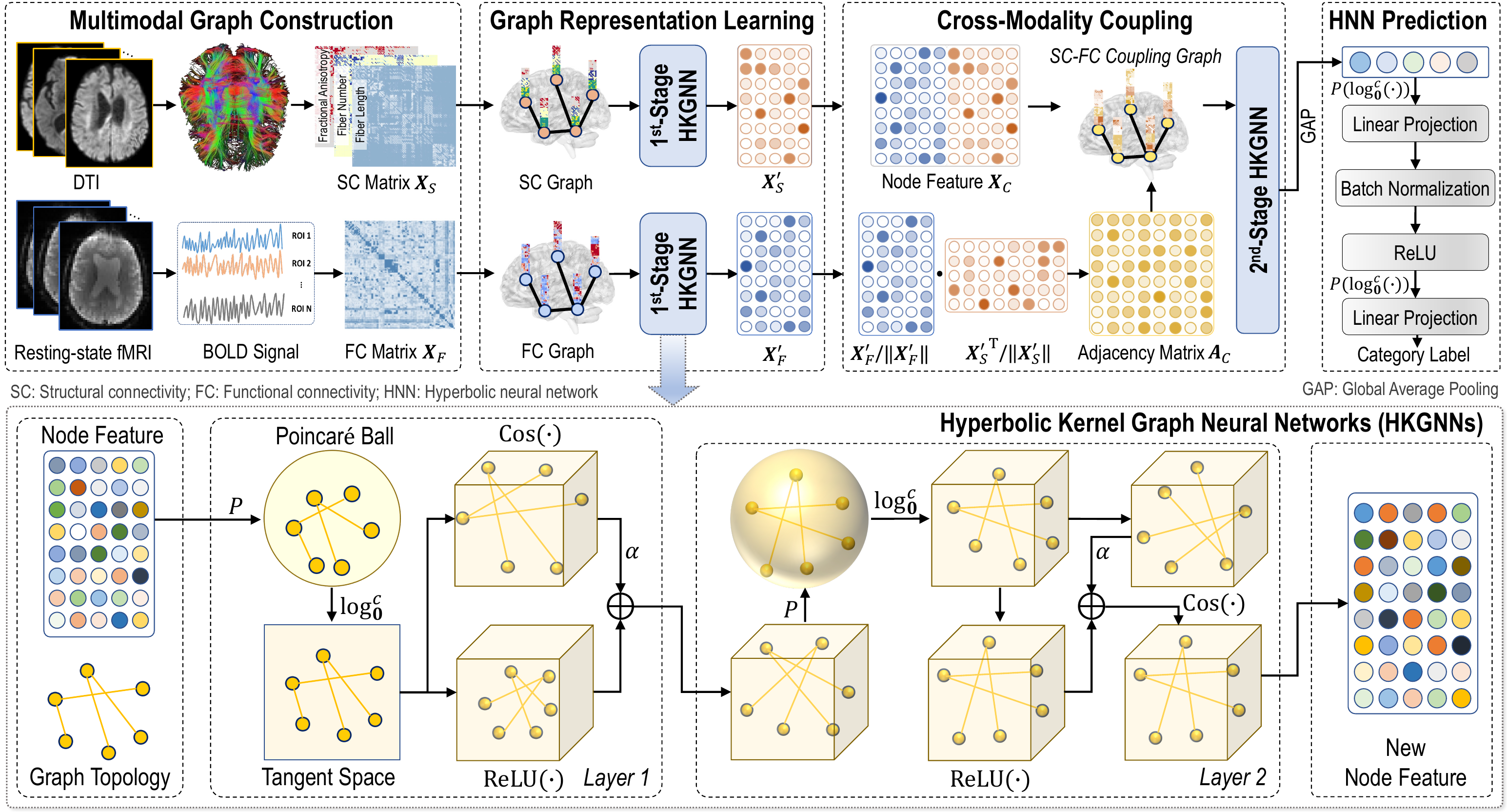}
\caption{Illustration of the proposed hyperbolic kernel graph fusion (HKGF) framework for neurocognitive decline analysis with multimodal data. 
Using DTI and fMRI input data as an example, this framework comprises four major components: (1) multimodal graph construction, (2) graph representation learning through a family of hyperbolic kernel graph neural networks (HKGNNs), 
(3) cross-modality coupling for feature fusion by capturing local to global connectivity interactions among brain regions, and (4) a prediction module using a new hyperbolic neural network (HNN).}
\label{fig_framework}
\end{figure*}

Extensive evidence has shown that the human brain is hierarchically organized in both structure and function, enabling efficient and flexible information processing at multiple levels~\cite{yeo2011organization,margulies2016situating,meunier2009hierarchical,betzel2014changes,zhou2006hierarchical,chen2008revealing,hilgetag2020hierarchy}. 
For instance, Yeo~\etal~\cite{yeo2011organization} demonstrated that the cerebral cortex can be divided into large-scale networks (\eg, default mode or frontoparietal) composed of anatomically distinct but functionally coherent regions. In addition, Margulies~\etal~\cite{margulies2016situating} revealed a major gradient of cortical organization, extending from unimodal to cross-modal regions, reflecting a functional hierarchy. 
\if false
On the structural side, studies by {\color{blue}Meunier~\etal~\cite{meunier2009hierarchical}}
and Betzel~\etal~\cite{betzel2014changes} 
\fi 
On the structural side, previous studies~\etal~\cite{betzel2014changes} have shown that the anatomical connectivity of the brain exhibits a modular and hierarchical topology, with smaller subnetworks embedded in larger integrated systems.  
\if false
Extensive evidence has shown that the human brain exhibits a hierarchical organization, spanning both structure and function brain connectivity networks~\cite{moreno2014hierarchical,zhou2006hierarchical,chen2008revealing,hilgetag2020hierarchy}.  
For instance, Yeo~\etal~\cite{yeo2011organization} studied a coarse parcellation that organized the human cerebral cortex into a 7 networks as as well a finer parcellation that identifies 17 networks, where were estimated to divide the 7 networks into smaller subnetworks. 
The parcellations yielded networks comprising primarily either anatomically contiguous regions (\eg, visual cortex) or regions that were broadly distributed across the cortical surface (\eg, heteromodal association cortex), suggesting the existence of multiple large-scale hierachnical networks that are interconnected throughout the association cortex. 
\fi 
Unfortunately, most current multimodal data fusion methods are formulated in Euclidean space, which makes them inadequate for capturing the inherently non-Euclidean  hierarchical structure of structural or functional  brain networks. 
As a result, their capacity to model complex interactions across different modalities is significantly limited. 
Hyperbolic space provides a promising solution to address this limitation 
because its negative curvature enables the volume to expand exponentially, which is suitable for modeling the hierarchical structure of brain networks~\cite{chami2019hgcn}.

Several previous studies have explored hyperbolic learning in medical data analysis. 
Zhang~\etal~\cite{zhang2016hyperbolic} developed a hyperbolic space sparse coding method for predicting Alzheimer's disease progression 
by mapping ventricular morphometry features mapped into hyperbolic space. 
Yu~\etal~\cite{yu2022skin} introduced a hyperbolic prototype network to leverage class hierarchy for skin lesion recognition, achieving improved classification performance. 
\if false
Qiao~\etal~\cite{qiao2024hyden} proposed 
a hyperbolic density embedding 
framework for aligning medical images and textual reports, demonstrating superior performance in zero-shot classification and retrieval tasks.
\fi 
Despite these advances, existing hyperbolic methods typically focus on single-modality data representation learning.  
Recently, Zhang~\etal~\cite{zhang2023multimodal} proposed a multimodal fusion method utilizing the hyperbolic graph convolutional neural network (HGCN)~\cite{chami2019hgcn} to integrate fMRI and DTI data, improving the identification of abnormal structural and functional disruptions associated with Alzheimer's disease.
However, due to the complex Riemannian operations in hyperbolic space, such as M\"obius addition and multiplication, these methods often have high computational complexity. 
Furthermore, existing methods often fail to explicitly capture local-to-global dependencies among structurally and functionally connected brain ROIs, which is crucial for the analysis of neurocognitive decline.

To this end, we propose a novel Hyperbolic Kernel Graph Fusion (HKGF) framework tailored for neurocognitive decline analysis using multimodal neuroimaging data. 
As illustrated in Fig.~\ref{fig_framework}, the proposed HKGF is composed of four key components: (1) a \emph{multimodal graph construction} module, (2) a \emph{graph representation learning module} that encodes structural and functional brain graphs in hyperbolic space 
through a general and flexible family of novel hyperbolic kernel graph neural networks (HKGNNs), 
(3) a \emph{cross-modality coupling module} that integrates multimodal brain graphs by explicitly capturing ROI-level dependencies within and across imaging modalities, 
and (4) a new \emph{hyperbolic neural network} (HNN) that facilitates downstream predictions. 
In particular, HKGNNs extend  
conventional GNNs 
by incorporating curvature-aware kernel functions in hyperbolic space, enabling the modeling of both local and global dependencies among brain regions while preserving the hierarchical structure of brain networks. Compared to conventional hyperbolic GNNs, our model avoids complex Riemannian operations by adopting an efficient kernel-based formulation, resulting in low computational cost. 
To the best of our knowledge, this is one of the first attempts to design and integrate hyperbolic kernels with graph neural networks for analyzing image-based neurocognitive decline. 
We also introduce a transfer learning strategy to reduce potential data scarcity by pretraining models on over 3,800 auxiliary fMRI scans. 
Extensive experiments 
on two target cohorts with 231 subjects show that HKGF outperforms state-of-the-art methods in multiple tasks. 
\if false
To address the small target data issue, we propose using a transfer learning strategy by pertaining models on over 3,800 auxiliary fMRI scans. 
Extensive experiments on 231 subjects with DTI and fMRI data from two target cohorts 
demonstrate the superiority of HKGF over state-of-the-art approaches in three tasks of neurocognitive decline prediction. 
\fi 
The source code and  pretrained models
can be accessed \href{https://github.com/mxliu/ACTION-Software-for-Functional-MRI-Analysis/tree/main}{online}.

The major contributions of this work are listed below. 
\begin{itemize}
\vspace{-1mm}
\item We propose a general Hyperbolic Kernel Graph Fusion (HKGF) framework for automated neurocognitive decline analysis by modeling the intrinsic hierarchical structure of brain networks and integrating multimodal neuroimaging data (\eg, DTI, fMRI, and ASL) into a unified end-to-end learning framework.

\item We design a family of hyperbolic kernel graph neural networks to represent structural/functional brain connectivity networks. 
Compared to conventional GNNs, our HKGNNs improve representation capacity by leveraging curvature-aware kernel functions to model complex hierarchical structures in brain networks. Compared to existing hyperbolic GNNs, HKGNNs significantly enhance computational efficiency by avoiding costly Riemannian operations. 

\item We introduce a cross-modality coupling module that explicitly captures interactions between heterogeneous neuroimaging modalities to facilitate effective multimodal fusion in hyperbolic space.

\item We conduct extensive experiments on multiple cohorts with multimodal data, 
demonstrating the superiority of HKGF over state-of-the-art Euclidean and non-Euclidean methods in multiple tasks. 

\end{itemize}

The remainder of this paper is structured as follows. Section~\ref{S2} reviews the most relevant work. Section~\ref{S3} describes the proposed method. Section~\ref{S4} outlines the experimental setup and presents the results. Section~\ref{S5} analyzes the impact of several key components. Section~\ref{S6} concludes this paper.

\section{Related Work}
\label{S2}

\subsection{Image-based
Neurocognitive Decline Analysis}


Neurocognitive impairment involves a range of cognitive deficits and is linked to abnormalities in brain structural and functional connectivity, as identified through imaging techniques like DTI and fMRI~\cite{oishi2011dti}. 
In particular, DTI quantifies white matter integrity and structural connectivity by modeling water diffusion along axonal tracts~\cite{basser1994mr}, while resting-state fMRI captures functional connectivity among ROIs by measuring correlated BOLD signal fluctuations. 
Oishi~\etal~\cite{oishi2011dti} applied DTI analysis to neurocognitive impairment and identified white matter degeneration in specific regions, particularly in limbic and association fibers, as potential biomarkers for early-stage diagnosis of Alzheimer's disease. 
Chen \etal~\cite{chen2023cerebral} used fMRI to examine neurocognitive impairment and observed FC alterations that reflect abnormalities in intrinsic brain networks.  
Liu \etal~\cite{ereira2024early} employed dynamic FCs derived from resting-state fMRI and proposed a GNN framework to improve the classification of mild cognitive impairment and Alzheimer's disease.
Wang~\etal~\cite{wang2024leveraging} used resting-state fMRI data for HIV-related neurocognitive impairment detection by incorporating brain community structure information for brain network analysis.

Multimodal data fusion has emerged as a powerful strategy to leverage the complementary information from diverse modalities such as DTI and fMRI, and ASL. 
Early efforts in multimodal brain imaging fusion primarily focus on feature-level concatenation of DTI and fMRI features~\cite{zhu2014fusing}. 
Beyond direct feature integration, Broser~\etal~\cite{broser2012functional} proposed an fMRI-guided probabilistic tractography framework to map cortico-cortical and cortico-subcortical language networks in children, integrating DTI-based connectivity with fMRI activation for individualized white matter analysis.
Iyer~\etal~\cite{iyer2013inferring}  proposed a DWI-guided Bayesian network structure learning approach that uses structural priors from DWI to infer directed functional networks from fMRI data. 
While this early work offered valuable insights into multimodal integration, it did not explicitly model the topological organization of brain connectivity.   
To address this limitation, recent studies explored graph-based representations to capture the topological structure of brain networks. 
Zhang \etal~\cite{zhang2023multi} developed a multimodal GNN framework that integrates brain networks from sMRI and PET via shared adjacency matrices and late fusion strategies, demonstrating its effectiveness in Alzheimer's disease classification.  
Bagheri~\etal~\cite{bagheri2023brain} proposed a Bayesian graph-based framework that integrates DTI-derived structural priors into causal discovery from fMRI data, improving the estimation of effective brain connectivity. 
These graph-based works are formulated in Euclidean geometry, which limits their ability to capture the hierarchical structure inherent in brain networks. 
To address this, Zhang~\etal~\cite{zhang2023multimodal} proposed a hyperbolic GCN framework that integrates DTI and fMRI data to enhance classification performance. However, the reliance on complex Riemannian operations in hyperbolic space results in high computational complexity.

\subsection{Brain Network Representation Learning} 

Existing studies often represent the brain as a functional or structural connectivity network derived from fMRI or DTI data. These networks are typically modeled as \emph{graphs}, where nodes denote regions-of-interest (ROIs) and edges encode functional or structural connections reflecting synchronized activity or anatomical linkages. 
To represent brain networks/graphs, researchers often extract node-level and graph-level features. 
Node-level features are computed for each ROI to assess the importance of individual brain regions, while graph-level features describe global network properties and inter-regional interactions, providing an overall view of the brain network.  

Learning-based methods have been developed for brain disorder analysis by feeding these predefined brain network features to machine learning algorithms for classification or regression. 
Recent efforts have focused on developing data-driven approaches for brain network analysis. 
For example, Kawahara~\etal~\citep{kawahara2017brainnetcnn} proposed BrainNetCNN, a CNN architecture tailored for structural brain network analysis using DTI-derived connectomes. It employs edge-to-edge, edge-to-node, and node-to-graph convolutions to capture topological features for neurodevelopmental outcome prediction.
Li~\etal~\cite{li2021braingnn} proposed BrainGNN to analyze resting-state fMRI data and identify disease-related neurological biomarkers. It takes brain networks as input and uses two node-level graph convolutional layers to learn representations that capture both topological and functional connectivity patterns. 
Recently, Chami~\etal~\cite{chami2019hgcn} proposed HGCN, which extends GCNs into hyperbolic space to better capture hierarchical structures in brain graphs. It embeds SC/FC networks using Möbius-based convolutions and generates graph-level embeddings via hyperbolic-to-Euclidean projection and pooling for classification.
However, these methods often struggle to explicitly capture the local-to-global dependencies between structurally and functionally connected brain ROIs, which are essential for understanding neurocognitive decline. 
In this work, we will develop a family of hyperbolic kernel GNN models aimed at capturing both local and global dependencies among brain regions while maintaining the hierarchical structure of brain networks. 
\if false
Hyperbolic graph neural networks (HGNNs) leverage hyperbolic geometry in graph neural networks to enhance the representational capacity of graph data.  
\fi

\section{Methodology}
\label{S3}

\begin{figure}[!t]
\setlength{\abovecaptionskip}{0pt}
\setlength{\belowcaptionskip}{0pt}
\setlength\abovedisplayskip{0pt}
\setlength\belowdisplayskip{0pt}
\centering
\includegraphics[width=0.49\textwidth]{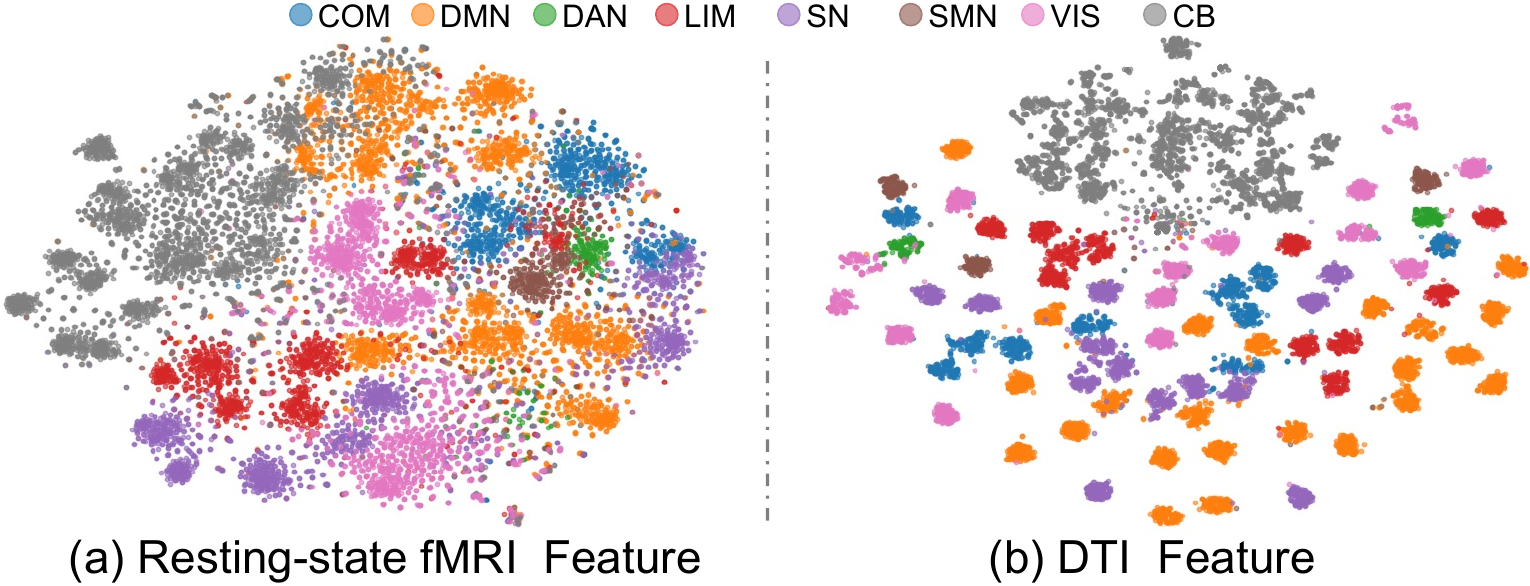}
\caption{T-SNE~\cite{van2008visualizing} visualizations of regional (a) fMRI and (b) DTI features for 
subjects from HAND cohort~\cite{wang2024leveraging}. 
For fMRI data, each brain ROI defined by AAL atlas is represented by a 116-dimensional vector, where each element corresponds to the functional connectivity (measured using Pearson correlation coefficients) with all other ROIs. 
For DTI data, each ROI is represented by a 348-dimensional feature vector, capturing its structural connectivity with other ROIs in terms of white matter fiber number (FN), fractional anisotropy (FA), and fiber length (FL) with other ROIs.  
Each point denotes an ROI for a specific subject, colored by its assigned group among the seven Yeo networks~\cite{yeo2011organization} or the cerebellum (CB). 
The seven Yeo networks include control (CON), default mode (DMN), dorsal attention (DAN), limbic (LIM), salience/ventral attention (SN), somatomotor (SMN), and visual (VIS) networks.
} 
\label{fig_hierachy}
\end{figure}

Previous evidence indicates that the human brain has a hierarchical organization across both structural and functional connectivity networks~\cite{yeo2011organization,margulies2016situating,meunier2009hierarchical,betzel2014changes}. 
To further explore this hierarchical structure in the context of neurocognitive decline, we visualize regional fMRI and DTI features from 137 subjects in an HIV-associated neurocognitive disorder (HAND) cohort~\cite{wang2024leveraging} using t-SNE~\cite{van2008visualizing}, as shown in Fig.~\ref{fig_hierachy}. 
Each of the 116 brain regions-of-interest (ROIs), defined by the Automated Anatomical Labeling (AAL) atlas~\cite{rolls2020automated}, is assigned to one of seven functional subnetworks identified by Yeo~\etal~\cite{yeo2011organization} or to the cerebellum (CB) group, based on maximum voxel overlap with the MNI152 template. 
ROIs that could not be reliably assigned to any Yeo functional network or the cerebellum (14 in total) based on voxel overlap were excluded from the t-SNE visualization.
 
These subnetworks include the control (CON), default mode (DMN), dorsal attention (DAN), limbic (LIM), salience/ventral attention (SN), somatomotor (SMN), and visual (VIS) networks. 
Each point in the figure represents a specific ROI from an individual subject. 
As shown in Fig.~\ref{fig_hierachy}~(a), brain regions within the same subnetwork display similar functional connectivity (FC) fingerprints, even when they are spatially separated across different cortical areas. 
A similar pattern is observed in Fig.~\ref{fig_hierachy}~(b) for DTI features. 
These consistent trends across modalities suggest that spatially distributed regions within each subnetwork work in coordination. 
This supports the view that the human brain exhibits a hierarchical organization in both structure and function, enabling efficient and flexible information processing across multiple levels. 
Motivated by these findings, we design a novel Hyperbolic Kernel Graph Fusion (HKGF) framework to explicitly capture the hierarchical organization of brain networks for neurocognitive decline analysis, with details introduced below.  

\subsection{Preliminaries} 
\noindent\textbf{Notations}. 
We use $\mathbb{R}^n$ and  $\mathbb{B}_c^n$ to denote the $n$-dimensional Euclidean space, $n$-dimensional Poincar\'e model with curvature $-c$, respectively. 
We omit the $n$ or $c$ when $n=1$ or $c=1$ for simplicity. 
The matrices, vectors and scalars are denoted by bold capital letters (e.g., $\bm{X}$), bold lower-case letters (e.g., $\bm{x}$) and thin letters (e.g., $x$), respectively.

\noindent\textbf{Hyperbolic Space}.  
 Hyperbolic space is a Riemannian manifold characterized by constant negative curvature~\cite{ratcliffe1994foundations}. Multiple isometric models have been proposed to represent hyperbolic space~\cite{cannon1997hyperbolic}.  Following~\cite{nickel2017poincare,ganea2018hyperbolic2,fang2023poincare}, 
 we adopt the Poincaré model in this work.  
The Poincaré ball $(\mathbb{B}^n_c, g_c^n)$ is an $n$-dimensional ball defined as:
\begin{equation}
\mathbb{B}^n_c = \{\bm{z} \in \mathbb{R}^n : c\|\bm{z}\|^2 < 1,\, c > 0\},
\end{equation}
with negative curvature $-c$ and the Riemannian metric $g_c^n$. 
As an analytic framework, gyrovector space theory facilitates operations in hyperbolic geometry~\cite{ungar2008analytic}, with the M\"obius gyrovector model well-aligned with the structure of the Poincaré ball~\cite{ungar1998pythagoras}. For example, for any two points $\bm{z}_i,\bm{z}_j\in\mathbb{B}_c^n$, the M\"obius addition is given by: 
\begin{equation}\label{eq:mobiusaddition}
\small
\bm{z}_i \oplus_c \bm{z}_j = \frac{(1 + 2c \langle \bm{z}_i, \bm{z}_j \rangle + c\|\bm{z}_j\|^2)\bm{z}_i + (1 - c\|\bm{z}_i\|^2)\bm{z}_j}{1 + 2c\langle \bm{z}_i, \bm{z}_j \rangle + c^2\|\bm{z}_i\|^2\|\bm{z}_j\|^2},
\end{equation}
where $\langle \bm{z}_i, \bm{z}_j \rangle$ denotes the Euclidean inner product. 
This non-Euclidean operation serves as the basis for defining hyperbolic geometric distances, which are computed as:
\begin{equation}\label{eq:Poincaredistance}
d_c^{\mathrm{H}}(\bm{z}_i, \bm{z}_j) = \frac{2}{\sqrt{c}} \tanh^{-1}(\sqrt{c}|\bm{z}_i \oplus_c (-\bm{z}_j)|).
\end{equation}
In addition to M\"obius operations within the manifold, computations in hyperbolic space often require projections to the tangent space (known as Euclidean space).  
This can be achieved by the following logarithmic map~\cite{ganea2018hyperbolic2}:  
 $\log_{\bm{0}}^c: \mathbb{B}_c^n \to T_{\bm{0}},~\mathbb{B}_c^n \cong \mathbb{R}^n$ for $\bm{z} \in \mathbb{B}_c^n \setminus \{\bm{0}\}$ is
\begin{equation}\label{eq:log0map}
\log_{\bm{0}}^c(\bm{z})
= \tanh^{-1}\!\bigl(\sqrt{c}\,\|\bm{z}\|\bigr)\,\frac{\bm{z}}{\sqrt{c}\,\|\bm{z}\|}.
\end{equation}

\subsection{Proposed Framework} 

As shown in Fig.~\ref{fig_framework}, the HKGF consists of a \emph{multimodal graph construction} module, a \emph{graph representation learning} module that encodes both structural and functional brain graphs in hyperbolic space using a general and flexible family of novel hyperbolic kernel graph neural networks ({HKGNNs}),
a \emph{cross-modality coupling} module that explicitly models region-of-interest (ROI) 
dependencies within and across imaging modalities, and a  \emph{hyperbolic neural network} (HNN) for downstream predictive tasks. 
In particular, the proposed HKGNNs extend traditional convolutional graph neural networks by integrating curvature-aware kernel functions within hyperbolic space, allowing for effective modeling of both local and global relationships among brain regions while maintaining the hierarchical nature of brain networks. 
Unlike conventional hyperbolic GNNs, our approach avoids computationally intensive Riemannian operations by adopting an efficient kernel-based design, significantly reducing computational cost and improving scalability.

\subsubsection{Multimodal Graph Construction}
Three modalities are used in this work, including DTI, resting-state fMRI, and ASL. 
Based on these modalities, we construct SC or FC networks for each subject.

\textbf{FMRI-based FC Graph Construction}.   
From resting-state fMRI, we will extract mean time series from $N$ ($N=116$ in this work) ROIs per subject, defined by the AAL atlas. 
With each ROI treated as a specific node, a fully-connected functional connectivity (FC) network is constructed by computing the Pearson correlation coefficients between fMRI time series of all ROI pairs, resulting in a {symmetric matrix $\bm{X}_F \in \mathbb{R}^{N \times D}$ ($D=N$ in this case)}. 
The original node feature for node $j$ is given by the $j$-th row of $\bm{X}_F$, and $\bm{X}_F$ is referred to as the node feature matrix. 
Since fully-connected brain networks may contain noisy or redundant connections, following~\cite{kim2021learning}, 
we empirically retain only the top $50\%$ of the strongest edges in each FC graph to construct a sparse binary adjacency matrix $A_{f} = (a_{ij}) \in \mathbb{R}^{N \times N}$, where $a_{ij}$ denotes the weight of an existing edge between two nodes, and $a_{ij} = 0$ otherwise.

\textbf{DTI-based SC Graph Construction}.  
For each subject, three complementary structural connectivity (SC) metrics are calculated from DTI data, including fiber number (FN), fractional anisotropy (FA), and fiber length (FL). 
These metrics are then used to construct a subject-specific SC graph. 
Each node in the graph corresponds to a brain ROI and is represented by the concatenated FN, FA, and FL features, covering both microstructural and macrostructural properties. 
{Denote the DTI-based node feature matrix as $\bm{X}_S \in \mathbb{R}^{N \times D}$ ($D=3N$ in this case),}  
where  
the original feature for the $j$-th node is represented by the $j$-th row of $\bm{X}_S$.  
Each edge in the SC graph is defined as the sum of the three metrics between the corresponding ROI pair, capturing the overall strength of structural connections.

\subsubsection{Graph Representation Learning with HKGNNs}
We aim to design a family of hyperbolic kernel graph neural networks (HKGNNs) to capture  hierarchical organization from brain structural and/or functional connectivity networks. 
In the following, we first introduce new hyperbolic kernels and then develop a new hyperbolic kernel graph convolutional network (HKGCN) and a hyperbolic kernel graph attention network (HKGAT). 
Throughout this paper, we refer to HKGF with the HKGCN backbone as \textbf{HKGF$_1$}, and HKGF with the HKGAT backbone as \textbf{HKGF$_2$}.

\textbf{Hyperbolic Kernels}. 
To effectively capture both global and local geometric structures in hyperbolic space, we introduce a hyperbolic arc-cos (HAC) kernel and a hyperbolic radial basis function (HRBF) kernel, inspired by their Euclidean counterparts (\ie, the arc-cos kernel~\cite{cho2009kernel} and the RBF kernel~\cite{rahimi2007random}). 
The $n$-th arc-cosine kernel is formulated as:
\begin{equation}
\scriptsize
k_n^{\mathrm{AC}}(\bm{x}_i,\bm{x}_j)=2 \int \Theta(\bm{w}^{\top}\bm{x}_i)(\bm{w}^{\top}\bm{x}_i)^n
\Theta(\bm{w}^{\top}\bm{x}_j)
(\bm{w}^{\top}\bm{x}_j)^n 
\frac{e^{-\frac{\|\bm{w}\|^2}{2}}}{(2\pi)^{d/2}} d(\bm{w}), 
\end{equation}
where $\bm{x}$$\in$ $\mathbb{R}^D$ and  $\Theta(x)$$=\frac{1}{2}[1+\mathrm{sign}(x)]$ denotes the Heaviside step function. 
{With $n=1$ and
an Rectified Linear Unit (ReLU) as the mapping function, the arc-cosine kernel~\cite{cho2009kernel} can be written as:
\begin{equation}
\small
 k^{\mathrm{AC}}(\bm{x}_i,\bm{x}_j) = 2\int \mathrm{ReLU}(\bm{w}^{\top}\bm{x}_i)\mathrm{ReLU}(\bm{w}^{\top}\bm{x}_j)\frac{e^{-\frac{\|\bm{w}\|^2}{2}}}{(2\pi)^{d/2}}d\bm{w}.
\end{equation}
where 
\[
     \small
\mathrm{ReLU}(y) = 
\begin{cases}
y, & \text{if } y \ge 0 \\
0, & \text{otherwise}
\end{cases}.
\]
In this way, $\Theta(\bm{w}^{\top}\bm{x}_i)(\bm{w}^{\top}\bm{x}_i)^{n=1}=\mathrm{ReLU}(\bm{w}^{\top}\bm{x}_i)$.
If ReLU is used as a mapping function in a standard neural network, 
this arc-cos kernel can be viewed as the inner product of the mapped features of two samples $\bm{x}_i$ and $\bm{x}_j$ in a specific layer of the neural network. 
}

Inspired by this kernel, our HAC kernel is defined as:  
\begin{equation}
k^{\mathrm{HAC}}(\bm{z}_i, \bm{z}_j) = 2 \int \phi^{\mathrm{HAC}}(\bm{z}_i, \bm{w}) \phi^{\mathrm{HAC}}(\bm{z}_j, \bm{w}) p(\bm{w}) d\bm{w},
\end{equation}
where $p(\bm{w})$ is the probability distribution function of $\bm{w}$,  $\phi^{\mathrm{HAC}}(\bm{z}, \bm{w}) = f(\bm{w}^\top \log_{\bm{0}}^c(\bm{z}) + b)$ is a non-linear transformation of $\log_{\bm{0}}^c(\bm{z})$,  with
$\log_{\bm{0}}^c(\cdot)$ denoting the logarithmic map from the Poincar\'e ball to its tangent space (see Eq.~\eqref{eq:log0map}). 
This formulation generalizes the arc-cosine kernel mapping $f^{\mathrm{AC}}(\bm{x})=\mathrm{ReLU}(\bm{w}^{\top}\bm{x}_i)$ to hyperbolic space, while allowing the use of arbitrary nonlinear functions $f(\cdot)$.  
When $f(\cdot)$ is distance-independent and operates on a shared tangent space representation, the HAC kernel aggregates information beyond local neighborhoods, thus 
\emph{help capture global similarity patterns in hyperbolic space}.

The HAC kernel can be approximated using {random features} methods~\cite{rudi2017generalization,bach2021learning},  
defined as:
\begin{equation}\label{eq:HAC_approx}
\Phi^{\mathrm{HAC}}(\bm{z}) = \sqrt{\frac{2}{M}} f(\bm{W}^\top \log_{\bm{0}}^c(\bm{z}) + \bm{b})
\end{equation}
where {\small{$\bm{W}= [\bm{w}_1, \cdots, \bm{w}_M]$}} denotes the collection of $M$ random vectors independently drawn from the distribution $p(\bm{w})$, and $M$ is the output feature dimension. 
The HAC kernel value is then approximated as an inner product of two data points in the mapped feature space:
\[
k^{\mathrm{HAC}}(\bm{z}_i, \bm{z}_j) \approx \langle  \Phi^{\mathrm{HAC}}(\bm{z}_i), \Phi^{\mathrm{HAC}}(\bm{z}_j)\rangle.
\]
{
Interpreting $f(\cdot)$ as a generic mapping function within a neural network layer, this approximation establishes a natural connection between the proposed hyperbolic kernel and deep learning frameworks.  
Accordingly, 
it can be seamlessly integrated into GNN models as a specialized mapping function, providing a principled way to incorporate hyperbolic geometry within graph-based learning frameworks. 
}

To capture local similarity among samples in hyperbolic space, we also introduce the HRBF kernel as follows:
\begin{equation}
\begin{split}
k^{\mathrm{HRBF}}(\bm{z}_i, \bm{z}_j)
= & \int e^{\mathrm{i} \bm{w}^\top (\log_{\bm{0}}^c(\bm{z}_i) - \log_{\bm{0}}^c(\bm{z}_j))} p(\bm{w}) d\bm{w},\\
= & \int \phi^{\mathrm{HRBF}}(\bm{z}_i) \phi^{\mathrm{HRBF}}(\bm{z}_j)p(\bm{w})d\bm{w}, 
\end{split}
\label{eq_KHRBF}
\end{equation}
where $\mathrm{i} = \sqrt{-1}$ denotes the imaginary unit, $p(\bm{w})$ is  a probability distribution over $\bm{w}$ and  $\log_{\bm{0}}^c(.)$ is the mapping function defined in Eq.~\eqref{eq:log0map}. 
Specifically, this HRBF kernel measures the similarity between a pair of mapped points $\log_{\bm{0}}^c(\bm{z}_i)$ and $\log_{\bm{0}}^c(\bm{z}_j)$ based on their Euclidean distance $d^{\mathrm{E}}$ 
 in the tangent space~\cite{rahimi2007random}: 
\begin{equation}
d^{\mathrm{E}}(\log_{\bm{0}}^c(\bm{z}_i), \log_{\bm{0}}^c(\bm{z}_j))=\|\log_{\bm{0}}^c(\bm{z}_i)-\log_{\bm{0}}^c(\bm{z}_j)\|. 
\end{equation}

According to the ``Curve Length Equivalence'' Theorem~\cite{fang2023poincare}, this  distance $d^E(\log_{\bm{0}}^c(\bm{z}_i),\log_{\bm{0}}^c(\bm{z}_j))$ in the tangent space actually approximates the hyperbolic geometric distance $d_c^{\mathrm{H}}(\bm{z}_i, \bm{z}_j)$ with a scaling factor.  
{
When this $d^E$ distance is small, the kernel value is high, indicating strong similarity between two samples. 
In contrast, as this distance increases, the kernel value tends to be $0$, indicating low similarity. 
With Eq.~\eqref{eq_KHRBF}, the HRBF kernel 
helps \emph{capture local structural information among samples in hyperbolic geometry}. 
}

Utilizing random Fourier features~\cite{rahimi2007random}, the HRBF kernel can be approximated using the inner product of the mapped features $\Phi^{\mathrm{HRBF}}(\bm{z}_i)$ and $\Phi^{\mathrm{HRBF}}(\bm{z}_j)$, formulated as:  
\begin{equation}\label{eq:HRBFapprox}
\Phi^{\mathrm{HRBF}}(\bm{z}) = \sqrt{\frac{2}{M}} \cos(\bm{W}^\top \log_{\bm{0}}^c(\bm{z}) + \bm{b}),
\end{equation}
{
By leveraging random feature approximation, the HRBF kernel can be flexibly applied in deep neural networks for downstream tasks. 
Similar to HAC, it can also be integrated into GNN models as a specialized mapping function.  
}

Both equations in Eq.~\eqref{eq:HAC_approx} and ~\eqref{eq:HRBFapprox} can be rewritten as
\begin{equation}\label{eq:unifiedMapping}
    \Phi^{H}(\bm{z})=\delta(\bm{W}^\top \log_{\bm{0}}^c(\bm{z}) + \bm{b}).
\end{equation}
where $\delta$ is the nonlinear activation. 
In this way, Eq.~\eqref{eq:unifiedMapping} can be viewed as a hyperbolic extension of an artificial neural network $\Phi(\bm{x})=\delta(\bm{W}^\top \bm{x} + \bm{b})$. 

\textbf{Proposed HKGCN}. 
For a brain SC/FC graph, we denote its adjacency matrix as $\bm{A}$ and its node feature matrix as $\bm{X} = [\bm{x}_1,\cdots,\bm{x}_N]^\top \in \mathbb{R}^{N \times D}$, where $ \bm{x}_i\in \mathbb{R}^{D}$ is the original feature of node $i$.
Given a projection function $P(\cdot)$, we integrate our HAC and HRBF kernels into the GCN~\cite{kipf2017gcn} framework
for graph feature learning, 
{
yielding a single-layer HKGCN.}  
\if false
Here,  {\small{$\bm{W}\in\mathbb{R}^{M\times M'}$}} is a trainable weight matrix, $\bm{b}\in\mathbb{R}^{M'\times 1}$ is a trainable bias vector, $\bm{A}\in N\times N$ means the  adjacency matrix, and $M'$ is the dimension of updated node features.
\fi 
Specifically, we first map the samples in Euclidean space to a Poincar\'e ball by:  
\begin{equation}
\small
\label{eq:project}
    \tilde{\bm{X}} = P(\bm{X}) =[P(\bm{x}_1),\cdots,P(\bm{x}_N)]^{\top},
\end{equation}
where {$P(\bm{x})=
\frac{\bm{x}}{\sqrt{c}\|\bm{x}\|}-\epsilon$ if $\sqrt{c}\|\bm{x}\|\ge1$}, and {$P(\bm{x})=\bm{x}$} otherwise, ensuring that all data points lie within the Poincar\'e ball. 
To capture the both local and global information in the brain network, we define the single layer HKGCN as: 
{
\begin{equation}\label{eq:HKGCNLayer}
\small
\begin{split}
\small    
    \Psi(\bm{A}, \bm{X}) =& f(\hat{\bm{A}}(\log_{\bm{0}}^c(\tilde{\bm{X}})\bm{W}+\bm{b}^{\top})) \\
    &+ \lambda\cos(\hat{\bm{A}}(\log_{\bm{0}}^c(\tilde{\bm{X}})\bm{W}+{\bm{b}}^{\top})),
\end{split}
\end{equation}
where $\bm{W}\in\mathbb{R}^{D\times M}$  
is a trainable weight matrix, $\bm{b}\in\mathbb{R}^{M}$ is a trainable bias vector}, $\hat{\bm{A}}\in\mathbb{R}^{ N\times N}$ is the normalized adjacency matrix, and $M$ is the dimension of updated node features. 
The hyperparameter $\lambda$ balances the contributions of the two kernels.

The first term in Eq.~\eqref{eq:HKGCNLayer} serves as a topology-aware approximation of HAC kernel for modeling \emph{global relationships}, where the nonlinear activation $f(\cdot)$ (ReLU in this work) is applied to node feature aggregation based on the normalized adjacency matrix. 
The second term 
serves as a topology-aware approximation of HRBF kernel for capturing local information, where the cosine function $\cos(\cdot)$ is applied for node feature aggregation. 
Instead of point-wise kernel activations in Eq.~\eqref{eq:HAC_approx} and Eq.~\eqref{eq:HRBFapprox}, HKGCN performs node feature aggregation in tangent space, followed by nonlinear kernel activations.  
This facilitates the seamless integration of hyperbolic kernels into GNN models for graph representation learning. 
We can construct a multi-layer HKGNN by stacking multiple HKGCN layers. 




\if false
The HKGCN-encoded SC and FC graph embeddings can be represented as 
    {{$\bm{X}_S' = \Psi(\bm{A}_S, \bm{X}_S)$}} and
    {{$\bm{X}_F' = \Psi(\bm{A}_F, \bm{X}_F)$}}, 
where $\Psi$ denotes {\color{blue}the mapping function } of the proposed 
multi-layer HKGCN model.  
\fi

\textbf{Proposed HKGAT}. 
Similar to HKGCN, we integrate the two hyperbolic kernels into the graph attention network ({GAT})~\cite{velivckovic2017graph} to define a single-layer HKGAT. 
In particular, HKGAT leverages an attention mechanism to learn node-specific aggregation weights, { allowing for flexible modeling of neighborhood contributions.}  
To this end, we first project the input feature matrix $\bm{X} = [\bm{x}_1,\cdots,\bm{x}_N]^\top \in \mathbb{R}^{N \times D}$ into a Poincar\'e ball using the project function defined in Eq.~\eqref{eq:project}. 
Let $\tilde{\bm{x}_i}=P(\bm{x}_i)$ and $\tilde{\bm{x}_j}=P(\bm{x}_j)$ denote the projected feature of node $i$ and node $j$, respectively. 
We first map 
$\tilde{\bm{x}}_i$ and $\tilde{\bm{x}}_j$ from the Poincar\'e ball to the tangent space using the logarithmic map $\log_{\bm{0}}^c(\cdot)$  defined in Eq.~\eqref{eq:log0map}. 
{
We then compute the 
attention score $e_{ij}$ as: 
   \begin{equation}
    \small
    \footnotesize
          e_{ij} =\mathrm{LeakyReLU}(\bm{a}^{\top}[(\bm{W} \log_{\bm{0}}^c(\tilde{\bm{x}}_i)+\bm{b})
          \|(\bm{W} \log_{\bm{0}}^c(\tilde{\bm{x}}_j)+\bm{b})]),
    \end{equation} 
where $\bm{W}\in\mathbb{R}^{M\times D}$ is a learnable weight matrix that transforms each feature vector into a shared embedding space}, 
and $\bm{a}\in\mathbb{R}^{2M}$ is a learnable attention vector. 
The operator $\|$ 
concatenates the transformed features of nodes $i$ and $j$ before applying the attention mechanism.  Besides, the LeakyReLU nonlinear function is defined as:
 \[
     \small
\mathrm{LeakyReLU}(y) = 
\begin{cases}
y, & \text{if } y \ge 0 \\
\beta y, & \text{otherwise} 
\end{cases},
\]
where $\beta \in (0,1)$ is a small positive constant that allows a small portion of negative input values to be preserved.
To obtain normalized attention coefficients, we apply the softmax function across all neighbors $\mathcal{N}(i)$ of node $i$:
\begin{equation}
\small
\alpha_{ij}=\frac{\exp(e_{ij})}{\sum_{k\in\mathcal{N}(i)}\exp(e_{ik})}.
\end{equation}
{Then, the new representation $\tilde{\bm{x}}'_i \in \mathbb{R}^{M}$ of node $i$ is obtained by aggregating the features of its neighbors, formulated as:
\begin{equation}\label{eq:1headHKGATLayer}
\small
    \tilde{\bm{x}}'_i = \sum\nolimits _{j\in\mathcal{N}(i)}\alpha_{ij} (\bm{W} \log_{\bm{0}}^c (\tilde{\bm{x}}_j)+\bm{b}),
\end{equation}
and the final output of the HKGAT layer is defined as:
\begin{equation}\label{eq_H_HKGAT}
\footnotesize
\begin{split}
\tilde{\bm{x}}''_i = &f(\tilde{\bm{x}}'_i) + \lambda\cos(\tilde{\bm{x}}'_i) \\
=& f(\sum\nolimits_{j\in\mathcal{N}(i)}\alpha_{ij}\bm{W}({\log_{\bm{0}}^c(\tilde{\bm{x}}_j)}+\bm{b}))  
\\
&+\lambda \cos(\sum\nolimits_{j\in\mathcal{N}(i)}\alpha_{ij}\bm{W}({\log_{\bm{0}}^c(\tilde{\bm{x}}_j)}+\bm{b})),
\end{split}
\end{equation}
where $\lambda$ is a scalar balancing the contribution of cosine activation.} 
With Eq.~\eqref{eq_H_HKGAT}, HKGAT is able to capture both local and global information across graph nodes.  
\if false
 The term $f(\cdot)$ corresponds to a specific nonlinear function in the HAC kernel~\eqref{eq:HAC_approx}, aiming to capture the global information across graph nodes.  
In contrast, the term $\cos(\cdot)$   draws motivation from the HRBF kernel, aiming to obtain the local relationships of the graph nodes. 
\fi 
Here, we use the Exponential Linear Unit (ELU) activation as a nonlinear function $f(\cdot)$, defined as: 

\[
\mathrm{ELU}(y) = 
\begin{cases}
y, & \text{if } y \geq 0 \\
\alpha(\exp(y) - 1), & \text{otherwise}
\end{cases},
\]
where $\beta \in (0,1)$ is a 
positive constant.  
{Here,  $\tilde{\bm{x}}''$ in Eq.~\eqref{eq_H_HKGAT} denotes the output of a single-head single-layer HKGAT.}

To capture complex relationships among nodes, we can use the multi-head attention technique (with $K$ heads) in each HKGAT layer, formulated as:
{
\begin{equation}
   {\hat{\bm{x}}}_i=\|_{k=1}^K \tilde{\bm{x}}_i^{''k}. 
\end{equation}}
By stacking single-head and/or multi-head HKGAT layers, we can obtain a multi-layer HKGAT. 

\if false
The HKGAT-encoded SC and FC graph embeddings can be represented as 
    {{$\bm{X}_S' = \Psi'(\bm{A}_S, \bm{X}_S)$}} and
    {{$\bm{X}_F' = {\Psi}'(\bm{A}_F, \bm{X}_F)$}}, 
where $\Psi^'$ denotes the mapping function of our multi-layer HKGAT.
\fi 

\if false
\textbf{Multi-Layer HKGNNs}. 
With a single HKGCN or HKGAT layer, we can construct a multi-layer HKGNN by stacking multiple HKGCN or HKGAT layers. 
{\color{red}
Taking HKGCN as an example, the HKGCN-encoded SC and FC graph embeddings can be represented as 
    {{$\bm{X}_S' = \Psi(\bm{A}_S, \bm{X}_S)$}} and
    {{$\bm{X}_F' = \Psi(\bm{A}_F, \bm{X}_F)$}}, 
where $\Psi$ denotes the mapping function of the proposed 
multi-layer HKGCN model. 
}    
\fi

The HKGNN-encoded SC and FC graph embeddings in Fig.~\ref{fig_framework} can be represented as {
    {{$\bm{X}_S' = \Psi(\bm{A}_S, \bm{X}_S)$}} and
    {{$\bm{X}_F' = \Psi(\bm{A}_F, \bm{X}_F)$}}, }
where $\Psi$ denotes the mapping function of the proposed 
multi-layer HKGCN or HKGAT.   
By integrating the two hyperbolic kernels, our multi-layer HKGCN and HKGAT  perform kernel-based feature transformations to effectively aggregate both local and global information from neighboring nodes at each layer. 
Their multi-layer architecture progressively capture local-to-global interactions in brain SC/FC networks, while the hyperbolic kernels inherently model their underlying hierarchical structure.


\subsubsection{Cross-Modality 
Coupling for Feature Fusion}
To capture global interactions between SC and FC graphs, we construct a novel \emph{SC-FC coupling graph} $G_C$ using an inner-product operation between the normalized embeddings of SC and FC graphs:
\begin{equation}
\small
\bm{A}_C = \frac{\bm{X}_F'}{\|\bm{X}_F'\|} \cdot \frac{{\bm{X}_S'}^\top}{\|\bm{X}_S'\|},
\end{equation}
where $\bm{A}_C \in \mathbb{R}^{N \times N}$ is the adjacency matrix representing global SC-FC interactions across ROIs. 
To define node features on this coupling graph, we concatenate the learned SC and FC representations to form the node feature matrix $\bm{X}_C = \big{[}\frac{\bm{X}_F'}{\|\bm{X}'_F\|}, \frac{\bm{X}_S'}{\|\bm{X}'_S\|}\big{]}$. 
Another HKGCN or HKGAT model 
is then applied to this coupling graph for generating fused features: $\bm{X}_{\text{CP}} = \Psi(\bm{A}_C, \bm{X}_C)$. 
This data-driven construction of the coupling graph enables explicit modeling of cross-modal interactions, in contrast to traditional methods (\eg, Pearson correlation~\cite{iyer2013inferring}) that rely on predefined assumptions. 
Moreover, the integration of HKGNNs allows for effective representation of the brain's hierarchical organization, capturing complex local-to-global dependencies. 

\if false 
To model global interactions between SC and FC graphs, 
we create a new \emph{SC-FC coupling graph} {\small{$G_C$}} 
through an inner-product operation between normalized SC and FC graph embeddings, defined as:
\begin{equation}
\bm{A}_C =\frac{\bm{X}_F'}{\|\bm{X}_F'\|} \cdot {\frac{\bm{X}'^{\top}_S}{\|\bm{X}_S'\|}},
\end{equation}
where {\small{$\bm{A}_C$}$\in \mathbb{R}^{N \times N}$} is the adjacency matrix of the SC-FC coupling graph, encoding global SC-FC interactions across ROIs. 
To represent each node/ROI in this coupling graph, we concatenate the learned SC and FC features for each node to obtain the new node feature {\small{$\bm{X}_C = [\bm{X}_S', \bm{X}_F']$}}. 
We then apply multi-layer HKGCN \(\Psi\)  to obtain fused representation: 
{\small{$ \bm{X}_{\text{CP}} = \Psi(\bm{A}_C, \bm{X}_C)$}}. 
By constructing the SC-FC coupling graph, we can explicitly capture SC-FC interactions. 
Compared to traditional statistical approaches (\eg, Pearson correlation~\cite{iyer2013inferring}) that rely on predefined assumptions, we construct a data-driven SC-FC coupling graph, 
while the integration of HKGCN and SC-FC coupling allows for effective modeling of hierarchical structures of brain networks~\cite{honey2009predicting,deco2014local}. 
\fi

{
\subsubsection{Prediction with Hyperbolic Neural Network}

Given the node feature matrix $\bm{X}_{\text{CP}} \in \mathbb{R}^{N \times M'}$} obtained from the SC-FC coupling module, we apply average pooling to obtain a compact vector representation $\bm{X}_H \in \mathbb{R}^{M'}$. 
To align with the hyperbolic nature of representations extracted by HKGCN/HKGAT, we design a new Hyperbolic Neural Network (HNN) for classification. 
Unlike conventional predictors that apply fully connected layers in Euclidean space, our HNN operates in the tangent space to preserve the geometric consistency and hierarchical properties~\cite{fang2023poincare} learned by HKGCN. Each layer applies this transformation:
\begin{equation}
\small
\bm{X}_{\text{HF}} = \delta\big( \bm{W}^{\top} \log_{\bm{0}}^c(P(\bm{X}_H)) + \bm{b} \big),
\end{equation}
where $\delta(\cdot)$ is the activation function (\ie, ReLU) and $P(\cdot)$ ensures that the input lies within the Poincar\'e ball. 
We stack two such layers to construct HNN. 
{
The final node feature $\bm{X}_{\text{HF}}$ is passed through a softmax layer for classification with cross-entropy loss. 
Compared to existing hyperbolic neural networks~\cite{ganea2018hyperbolic2,chami2019hgcn}, our HNN reduces computational complexity by avoiding costly operations in hyperbolic space. 

\if false 
{\color{red}
Given the feature $\bm{X}_{\text{CP}}\in\mathbb{R}^{N\times D}$ extracted from the SC-FC coupling module, we average it into a vector $\bm{X}_H\in\mathbb{R}^{64}$. 
To align with the hyperbolic nature of HKGCN-extracted representations, we design HNN for classification. 
Unlike conventional predictors that apply fully connected layers directly to GCN-extracted features, HNN operates in the tangent space to maintain geometric consistency and hierarchical properties~\cite{fang2023poincare} learned by HKGCN. 
In each layer, we apply the following transformation:   
\begin{equation}
\bm{X}_{\text{HF}} = \delta\big{(} \bm{W}^{\top} \log_{\bm{0}}^c(P(\bm{X}_H)) + \bm{b} \big{)}, 
\end{equation} 
where $\delta(\cdot)$ is the activation function (ReLU), 
and {\small{$P(.)$}} ensures the input remains in the Poincar\'e ball. 
We stack two layers of this transformation. 
The final representation \(\bm{X}_{\text{HF}}\) is then fed into a softmax layer for classification 
with a cross-entropy loss. 
Compared to existing hyperbolic networks~\cite{ganea2018hyperbolic2}, our HNN decreases the computational cost by avoiding complex transformations in hyperbolic space. 
\fi

\subsection{Implementation Details}   
{In the proposed HKGF$_1$, the HKGCN backbone consists of two hyperbolic kernel layers, both equipped with a hidden dimension of $64$ and the rescaling factor $\lambda = 0.01$. 
We use the first-stage HKGCN for SC/FC graph feature learning, and the second-stage HKGCN for SC-FC coupling graph representation learning. 
Similarly, the HKGF$_2$ employs an HKGAT backbone, also comprising two hyperbolic kernel layers with a hidden dimension of $64$  and a rescaling factor of $\lambda = 0.01$. 
The first layer of HKGAT uses $K=4$ attention heads, and the second layer uses a single head. 
We also use two-stage HKGAT modules in HKGF$_2$. 
} 
{Both HKGCN and HKGAT utilize the curvature parameter $c$ of 0.001}.} 
The HNN contains two layers (each with a hidden dimension of $M=32$). 
The Adam optimizer is used with a learning rate of {$0.0001$} and a weight decay of $1\times10^{-4}$. 
The batch size is $128$, and the training epoch is 50. 
To support reproducible research, we have made our source code publicly available via GitHub (see \href{https://github.com/mxliu/ACTION-Software/tree/main/HKGCN}{HKGCN} and \href{https://github.com/mxliu/ACTION-Software/tree/main/HKGAT}{HKGAT}).

\section{Experiment}
\label{S4}

\begin{table}[!tbp]
\setlength{\abovecaptionskip}{0pt}
\setlength{\belowcaptionskip}{0pt}
\setlength{\abovedisplayskip}{0pt}
\setlength{\belowdisplayskip}{0pt}
\renewcommand\arraystretch{0.6}
\centering
\caption{Demographic information of subjects from two target cohorts (target domains) and three auxiliary datasets (source domains). Age is presented as the mean$\pm$standard deviation. 
 ANI: asymptomatic neurocognitive impairment with HIV; 
 CN: control normal; 
 SMC: significant memory concern; 
 ASD: autism spectrum disorder; 
 MDD: major depressive disorder; 
 ADHD: attention-deficit/hyperactivity disorder; 
  M: Male; F: Female.}
	\scriptsize
	\centering
	\setlength{\tabcolsep}{0.5pt}
	    \label{demographic}
		\begin{tabular*}{0.48\textwidth}{@{\extracolsep{\fill}} l|ccccccc }
		\toprule
			~Dataset & Modality & Domain & Category   &Subject~\# & Sex~(M/F) & Age~ \\

            \midrule
        			\multirow{2}{*} {~ADNI}
		    &\multirow{2}{*}{fMRI,DTI} &\multirow{2}{*}{Target}    &SMC    &$46$	   &$16/30$	 &${75.63}_{5.43}$ \\
		    &&  &CN  &$48$   &$18/30$   &${75.02}_{4.04}$ \\
\midrule
				\multirow{2}{*} {~HAND}
		    &\multirow{2}{*}{fMRI,DTI}   &\multirow{2}{*}{Target} &ANI    &$68$	   &$68$/$0$	 &${33.07}_{6.18}$ \\
    			&&  &CN    &$69$   &$69$/$0$	 &${33.33}_{5.37}$ \\  
         
        \midrule   
			\multirow{2}{*} {~ABIDE}
	  	    &\multirow{2}{*}{fMRI} &\multirow{2}{*}{Source}    &ASD  &$351$	   &$308$/$43$	 &${16.90}_{8.00}$ \\
			&\multirow{2}{*}        &&CN   &$370$   &$295$/$75$	 &${16.60}_{6.80}$ \\   

        \midrule   
			\multirow{2}{*} {~MDD}
	  	    &\multirow{2}{*}{fMRI} &\multirow{2}{*}{Source}    &MDD  &$1,163$	   &$415$/$748$	 &${36.90}_{14.90}$ \\
			&\multirow{2}{*}        &&CN   &$1,104$   &$425$/$579$	 &${37.00}_{16.00}$ \\  
            
        \midrule   
			\multirow{2}{*} {~ADHD-200}
	  	    &\multirow{2}{*}{fMRI} &\multirow{2}{*}{Source}    &ADHD  &$285$	   &$232$/$52$	 &${12.20}_{3.10}$ \\
			&\multirow{2}{*}        &&CN   &$379$   &$196$/$183$	 &${13.20}_{3.50}$ \\              
			\bottomrule
	    \end{tabular*}
\label{tab_demographics}
\end{table}

\subsection{Materials and Data Preprocessing}
\subsubsection{Materials}
Both the Alzheimer's Disease Neuroimaging Initiative (ADNI) dataset~\cite{jack2008alzheimer} and an HIV-associated neurocognitive disorder (HAND) dataset~\cite{wang2024leveraging} are used as the target domains in this work. 
On ADNI, we aim to identify patients with significant memory complaints (SMC) from CN subjects, including 
paired resting-state fMRI and DTI data from 46 SMC subjects and 58  age- and sex-matched CN subjects. 
On HAND, we aim to identify asymptomatic neurocognitive impairment with HIV (ANI) from control normal (CN) subjects. 
Both resting-state fMRI and DTI data in HAND are collected from $68$ ANIs 
and age-matched $69$ HCs. 
Given the limited sample sizes in the two target domains, we employ a transfer learning strategy by pretraining deep learning models on large-scale auxiliary source domain data.  
These source domain data (with 3,806 resting-state fMRI scans) are from three public datasets, including ABIDE~\cite{di2014autism}, REST-meta-MDD Consortium~\cite{yan2019reduced}, and ADHD-200~\cite{adhd2012adhd}. 
Demographic information of the studied subjects from the source and target domains 
is reported in Table~\ref{tab_demographics}. 

\begin{table*}[!tbp]
\setlength{\abovecaptionskip}{0pt}
\setlength{\belowcaptionskip}{0pt}
\setlength{\abovedisplayskip}{0pt}
\setlength{\belowdisplayskip}{0pt}
    \centering
        \caption{Results (\%) of 
        different methods in SMC vs. CN classification on ADNI with fMRI and DTI data. 
        `$^*$' denotes that a competing method and HKGF$_1$ are significantly different ($p<0.05$ via $t$-test), while `$^\dag$' denotes that a competing method and HKGF$_2$ are significantly different.
        }
\renewcommand{\arraystretch}{0.5}
    \setlength\tabcolsep{1.5pt}{\scalebox{1}{
    \begin{tabular}{l|ccccccc|c|c}
        \toprule

          Method & AUC & ACC & F1 & BAC& SEN & SPE & PRE &$p$-value$^*$ &$p$-value$^\dag$  \\
\midrule 
          SVM~\cite{pisner2020support} & $55.08\pm{5.90}$ & $57.63\pm{2.60}$ & $59.92\pm{2.05}$ & $58.51\pm{1.25}$ & $59.19\pm{3.97}$ & $57.83\pm{5.54}$ & $64.42\pm{2.09}$  & $<0.001^*$ &  $<0.001^\dag$\\
         RF~\cite{breiman2001random} & $67.75\pm{3.54}$ & $60.42\pm{2.99}$ & $67.29\pm{2.42}$ & $60.80\pm{2.58}$ & $76.52\pm{3.68}$ & $45.08\pm{7.38}$ & $63.28\pm{2.39}$  & $<0.001^*$ & $<0.001^\dag$\\
       XGBoost~\cite{chen2016xgboost} & $70.54\pm{4.87}$ & $63.53\pm{3.87}$ & $67.20\pm{4.43}$ & $64.26\pm{4.74}$ & $70.42\pm{3.50}$&$58.10\pm{7.11}$ & $67.49\pm{4.75}$   &  $<0.001^*$  &  $<0.001^\dag$ \\
         \cmidrule{1-10}                   
    GCN-EF~\cite{kipf2017gcn} & $74.72\pm{4.31}$ & $73.65\pm{4.98}$ & $73.65\pm{6.22}$ & $75.41\pm{3.37}$ & $71.89\pm{10.27}$ & $78.93\pm{10.15}$ & $\bm{83.19}\pm{5.95}$  &  $<0.001^*$  &  $<0.001^\dag$\\
    {GCN-LF}~\cite{kipf2017gcn}&$70.02\pm{2.08}$& $72.22\pm{1.92}$ &$74.92\pm{2.49}$&$71.48\pm{1.78}$&$78.33\pm{4.59}$&$64.63\pm{5.20}$& $74.73\pm{2.46}$  & $<0.001^*$
  & $<0.001^\dag$\\

    GAT-EF~\cite{velivckovic2017graph} & $73.87\pm{1.47}$  & $75.84\pm{2.34}$  & $76.06\pm{3.16}$ & $75.62\pm{1.78}$ & $74.44\pm{7.55}$ & $76.81\pm{7.15}$ &  $80.86\pm{2.61}$  & $<0.001^*$ &  $<0.001^\dag$\\
    GAT-LF~\cite{velivckovic2017graph}&$77.67\pm{2.80}$&$75.68\pm{2.98}$& $77.48\pm{3.24}$ &$75.43\pm{2.60}$& $78.53\pm{6.14}$ & $72.32\pm{5.15}$ & $79.10\pm{1.55}$  & $0.330$  & $<0.001^\dag$\\  

    Transformer-EF~\cite{vaswani2017attention}&$66.95\pm{2.26}$&$68.64\pm{3.96}$&${67.83}\pm{2.53}$&$69.36\pm{4.09}$&$65.80\pm{6.51}$&$72.92\pm{13.80}$& $79.61\pm{9.27}$  & $0.001^*$   & $<0.001^\dag$\\
    
    {Transformer-LF}~\cite{vaswani2017attention}& $70.89\pm{7.81}$ &$71.66\pm{8.10}$&$75.26\pm{6.85}$&$71.02\pm{7.85}$&$81.76\pm{10.43}$&$60.29\pm{17.73}$&$74.99\pm{8.62}$  & $<0.001^*$  & $0.048^\dag$ \\

    GraphSAGE-EF~\cite{hamilton2017inductive}  & $73.82\pm{3.16}$ & $69.05\pm{1.62}$ & $69.56\pm{2.48}$ & $70.93\pm{3.10}$ & $67.67\pm{5.18}$ & $74.19\pm{8.02}$ & $77.38\pm{4.51}$  & $<0.001^*$ & $<0.001^\dag$ \\
    GraphSAGE-LF~\cite{hamilton2017inductive}  & $75.70\pm{3.77}$ & $72.55\pm{2.49}$ & $72.62\pm{4.00}$ & $74.34\pm{2.51}$ & $69.37\pm{6.21}$ & $\bm{79.32}\pm{3.82}$ & $81.70\pm{3.04}$  & $0.012^*$ & $<0.001^\dag$ \\   

    GIN-EF~\cite{xu2018gin} &$74.17\pm{4.10}$&$70.96\pm{3.14}$&$71.37\pm{3.43}$&$72.86\pm{2.46}$&$68.38\pm{4.31}$&$77.33\pm{3.44}$&$79.27\pm{3.22}$ & $<0.001^*$  & $<0.001^\dag$\\      
       
    GIN-LF~\cite{xu2018gin}  & $70.32\pm{2.34}$ & $69.42\pm{1.41}$ & $67.10\pm{3.04}$ & $69.88\pm{1.66}$ & $63.39\pm{6.81}$ & $76.37\pm{8.36}$ & $79.55\pm{3.86}$  & $<0.001^*$ & $<0.001^\dag$ \\

    {BrainNetCNN-EF}~\cite{kawahara2017brainnetcnn}  & $68.72\pm{8.60}$ & $61.85\pm{3.43}$ & $57.85\pm{5.58}$ & $62.92\pm{4.22}$ & $57.15\pm{9.20}$ & $68.70\pm{15.94}$ & $66.55\pm{11.12}$  & $<0.001^*$  &  $<0.001^\dag$ \\
    BrainNetCNN-LF~\cite{kawahara2017brainnetcnn} & $71.95\pm{6.53}$ & $74.07\pm{4.55}$& $75.25\pm{6.36}$&$75.74\pm{5.01}$&$75.66\pm{12.75}$&$75.82\pm{8.53}$& $80.80\pm{2.79}$  &  $<0.001^*$ & $<0.001^\dag$ \\   
    
    BrainGNN-EF~\cite{li2021braingnn} &$62.61\pm{6.15}$&$67.44\pm{6.59}$&$66.24\pm{11.18}$&$68.30\pm{3.22}$&$65.22\pm{14.36}$&$71.38\pm{8.04}$& $77.19\pm{3.96}$  & $0.004^{*}$ & $0.006^{\dag}$ \\ 
    BrainGNN-LF~\cite{li2021braingnn} &$72.21\pm{6.89}$&$71.02\pm{5.44}$&$73.75\pm{4.23}$&$72.40\pm{4.86}$&$76.59\pm{4.12}$&$68.22\pm{8.23}$&$75.22\pm{4.58}$   &  $0.023^*$ & $0.003^{\dag}$\\ 
    
    HGCN-EF~\cite{chami2019hgcn}  & $71.10\pm{4.04}$ & $69.28\pm{2.52}$ & $68.58\pm{4.29}$ & $69.79\pm{2.39}$ & $65.00\pm{3.12}$ & $74.58\pm{3.28}$ & $77.70\pm{2.51}$  & $<0.001^*$  & $<0.001^\dag$\\    
    HGCN-LF~\cite{chami2019hgcn} &$69.81\pm{5.59}$&$69.58\pm{3.90}$&$70.37\pm{3.58}$&$69.91\pm{4.43}$&$67.33\pm{5.22}$&$72.49\pm{11.04}$&$78.24\pm{7.50}$  & $<0.001^*$ & $<0.001^\dag$\\
    
    \cmidrule{1-10}
    HKGF$_1$~(Ours)   & ${76.22}\pm{2.31}$ & ${76.30}\pm{1.99}$ & ${77.59}\pm{2.54}$ &${78.10}\pm{2.13}$ &${78.34}\pm{4.77}$ & ${77.87}\pm{3.15}$ & ${81.32}\pm{1.99}$ & --  &-- \\
    HKGF$_2$~(Ours) & $\bm{80.42}\pm{2.25}$ & $\bm{81.26}\pm{2.62}$ & $\bm{82.65}\pm{2.64}$ &$\bm{80.58}\pm{2.11}$ &$\bm{84.81}\pm{4.08}$ & ${76.35}\pm{3.97}$ & ${81.75}\pm{3.11}$ &  -- & --\\
        \bottomrule
    \end{tabular}}}
    \label{tab_SMCfMRIDTI}
\end{table*}
\if false
\begin{table*}[!tbp]
\setlength{\abovecaptionskip}{0pt}
\setlength{\belowcaptionskip}{0pt}
\setlength{\abovedisplayskip}{0pt}
\setlength{\belowdisplayskip}{0pt}
    \centering
        \caption{Results (\%) of 
        different methods in ANI vs. CN classification on HAND with fMRI and DTI data. `$^*$' denotes that a competing method and HKGF$_1$ are significantly different ($p<0.05$ via $t$-test), while `$^\dag$' denotes that a competing method and HKGF$_2$ are significantly different. {\color{red}UPDATE!}
}
\renewcommand{\arraystretch}{0.5}
    \setlength\tabcolsep{1.5pt}{\scalebox{1}{
    \begin{tabular}{l|ccccccc|c|c}
        \toprule

          Method & AUC & ACC & F1 & BAC& SEN & SPE & PRE &$p$-value &$p$-value \\
\midrule 
          SVM~\cite{pisner2020support} & $51.96\pm{3.25}$ &$50.08\pm{3.01}$   & $48.70\pm{3.99}$  & $51.57\pm{3.55}$  & $47.50\pm{4.61}$  & $55.63\pm{4.05}$ & $52.58\pm{4.40}$ & $<0.001^*$ & $<0.001^\dag$ \\
         RF~\cite{breiman2001random}& $54.77\pm{3.05}$ & $49.25\pm{3.05}$ & $48.49\pm{3.49}$ & $50.63\pm{3.66}$ & $49.75\pm{5.80}$ & $51.49\pm{9.83}$ & $50.88\pm{3.82}$  & $<0.001^*$ & $<0.001^\dag$ \\
       XGBoost~\cite{chen2016xgboost} & $57.49\pm{6.34}$ & $54.61\pm{5.21}$ & $54.53\pm{5.76}$ & $54.97\pm{5.01}$ & $55.70\pm{7.13}$ & $54.23\pm{5.06}$ & $55.62\pm{5.54}$  &  $<0.001^*$ &  $<0.001^\dag$\\
         \cmidrule{1-10}                  
       GCN-EF~\cite{kipf2017gcn} & $65.24\pm{5.59}$ & $64.83\pm{3.36}$  & $61.97\pm{6.88}$ & $65.54\pm{4.08}$ & $64.74\pm{10.14}$ & $66.34\pm{10.34}$  & $69.09\pm{5.05}$  & {\color{red}{$0.039^*$}}  & $0.028^\dag$ \\   
       GCN-LF~\cite{kipf2017gcn} & $61.06\pm{2.71}$ & $63.16\pm{1.46}$  & $62.88\pm{2.66}$ & $63.35\pm{1.45}$ & $67.44\pm{6.54}$ & $59.27\pm{7.35}$  & $65.52\pm{4.04}$  & $<0.001^*$ & {\color{blue}{$0.011^\dag$}} \\       
      GAT-EF~\cite{velivckovic2017graph} & $65.10\pm{5.17}$ & $64.79\pm{3.07}$ & $61.91\pm{7.13}$  & $64.38\pm{3.19}$ & $65.60\pm{9.61}$ & $63.15\pm{6.56}$ & $67.63\pm{3.08}$  & {\color{red}{$0.019^*$}}  & $0.005^\dag$\\
       GAT-LF~\cite{velivckovic2017graph} & $62.61\pm{3.78}$ & $64.23\pm{3.49}$ & $66.48\pm{5.90}$  & $60.77\pm{5.75}$ & 
       $64.21\pm{2.42}$ &$67.94\pm{11.26}$ &  $67.58\pm{4.33}$  & $0.046^*$ & $<0.001^\dag$\\

    Transformer-EF~\cite{vaswani2017attention} & $54.92\pm{4.36}$ & $59.13\pm{2.69}$ & $57.97\pm{5.95}$ & $59.46\pm{1.52}$ & $64.56\pm{9.28}$ & $54.36\pm{10.01}$ & $57.90\pm{8.06}$  & $<0.001^*$ & $<0.001^\dag$ \\
     Transformer-LF~\cite{vaswani2017attention} & $55.49\pm{3.88}$ & $60.56\pm{3.81}$ & $53.86\pm{2.31}$ & $59.54\pm{2.00}$ & $64.81\pm{9.42}$ & $64.26\pm{12.69}$ &  $63.29\pm{5.59}$ & $<0.001^*$ & $<0.001^\dag$ \\
        
         GraphSAGE-EF~\cite{hamilton2017inductive} & $63.11\pm3.63$ & $64.23\pm{3.98}$ & $64.24\pm{6.10}$ & $64.21\pm{4.23}$ & $71.11\pm{10.18}$ & $57.32\pm{4.89}$ & $64.04\pm{3.10}$  & {\color{blue}{$0.025^*$}}  & $0.032^\dag$\\
        GraphSAGE-LF~\cite{hamilton2017inductive} & $62.31\pm{2.30}$ & $64.23\pm{1.90}$  & $62.70\pm{2.58}$ & $63.60\pm{1.46}$ & $64.01\pm{5.54}$ & $63.18\pm{5.55}$ & $64.21\pm{1.85}$  & $0.048^*$  &  {\color{blue}{$0.024^\dag$}}\\

          GIN-EF~\cite{xu2018gin}  & $62.90\pm{5.34}$ & $65.12\pm{4.34}$  & $62.51\pm{8.48}$ & $64.54\pm{4.62}$ & $66.07\pm{10.43}$ & $63.00\pm{9.74}$ & $68.73\pm{9.00}$  & {\color{red}{$0.022^*$}}  & {\color{blue}{$0.033^\dag$}}\\

          GIN-LF~\cite{xu2018gin}  & $ 61.30\pm{2.42}$ & $63.63\pm{2.80}$  & $61.05\pm{3.60}$ & $63.74\pm{1.94}$ & $64.91\pm{7.98}$ & $62.56\pm{11.56}$ & $69.01\pm{8.68}$  & $0.038^*$  & $<0.001^\dag$ \\
   
       BrainNetCNN-EF~\cite{kawahara2017brainnetcnn}  & $54.73\pm{4.46}$ & 
       $58.25\pm{1.83}$ & $57.97\pm{4.93}$ & $58.20\pm{1.96}$ & $63.12\pm{13.76}$ & $53.29\pm{16.81}$& $59.33\pm{4.92}$ & $<0.001^*$ 
       & $<0.001^\dag$\\
    BrainNetCNN-LF~\cite{kawahara2017brainnetcnn}  & $54.98\pm{5.88}$ & 
       $62.05\pm{4.76}$ & $57.79\pm{7.47}$ & $62.01\pm{4.94}$ & $61.08\pm{13.80}$ & $62.95\pm{13.16}$& $63.76\pm{11.95}$ & $<0.001^*$ & $<0.001^\dag$\\
            \textcolor{red}{BrainGNN-EF}~\cite{li2021braingnn} & $50.52\pm{0.79}$ & $51.44\pm{1.66}$ & $50.53\pm{6.14}$ & $51.40\pm{0.29}$ & $57.78\pm{12.79}$ & $45.05\pm{12.48}$ & $52.17\pm{0.57}$  &  & \\
            \textcolor{red}{BrainGNN-LF}~\cite{li2021braingnn} & $50.52\pm{0.79}$ & $51.44\pm{1.66}$ & $50.53\pm{6.14}$ & $51.40\pm{0.29}$ & $57.78\pm{12.79}$ & $45.05\pm{12.48}$ & $52.17\pm{0.57}$  &  & \\
        HGCN-EF~\cite{chami2019hgcn}  & $64.08\pm{5.42}$ & $63.66\pm{3.95}$ & $65.18\pm{7.17}$ & $63.81\pm{3.81}$ & $72.35\pm{10.23}$ & $55.27\pm{4.29}$ & $64.64\pm2.18$  &  $<0.001^*$ & $<0.001^\dag$ \\ 
        HGCN-LF~\cite{chami2019hgcn}  & $60.68\pm{1.58}$ & $65.12\pm{1.15}$ & $61.02\pm{4.13}$ & $65.04\pm{1.26}$ & $59.52\pm{8.43}$ & $70.57\pm{7.10}$ & $69.74\pm{5.21}$  & $<0.001^*$  & $<0.001^\dag$\\ 
          \cmidrule{1-10}
   HKGF$_1$~(Ours) & ${68.83}\pm{2.55}$ & ${71.53}\pm{2.59}$ & $\bm{69.72}\pm{5.15}$ &${71.54}\pm{2.53}$ & $\bm{71.30}\pm{8.73}$ & ${71.78}\pm{4.18}$ & ${73.62}\pm{1.72}$ & --  &-- \\
        HKGF$_2$ (Ours)& $\bm{69.74}\pm{1.77}$ & $\bm{71.86}\pm{2.39}$ & ${68.73}\pm{2.54}$ &$\bm{72.36}\pm{1.40}$ &${70.45}\pm{7.57}$ & $\bm{74.27}\pm{7.64}$ & $\bm{76.45}\pm{4.99}$ &--   & --\\
        \bottomrule
    \end{tabular}}}
    \label{tab_ANIfMRIDTI}
\end{table*}
\fi

\subsubsection{Data Preprocessing}
\label{S4_preprocessing}
All resting-state fMRI data were preprocessed using a popular pipeline, 
including  
magnetization stabilization, slice time correction, head motion correction, regression of confounding covariates such as white matter signal, ventricular signal, and head motion parameters, normalization to the Montreal Neurological Institute (MNI) space, spatial smoothing, and bandpass filtering. 
We extracted mean time series (\ie, blood-oxygen-level dependent signals) from 116 ROIs per subject, which were defined by the Automated Anatomical Labeling (AAL) atlas. 
Based on regional BOLD signals, we construct an FC network for each subject. 
All DTI data were preprocessed using 
an established pipeline, including brain tissue extraction, bias field correction, eddy current and head motion correction,
diffusion gradient direction adjusting, registering anatomical images of each subject to the diffusion space, 
and brain ROI partition based on AAL. 
For each subject, three SC metrics were calculated: fiber number (FN), fractional anisotropy (FA), and fiber length (FL). 
These metrics are concatenated to represent each ROI.  
The weight of each edge between a pair of ROIs is computed as the sum of the three metrics.  

\if false 
Multiple neuroimaging data will be preprocessed using established pipelines, with details introduced in the following. 

\textbf{Resting-State fMRI Preprocessing}. 
All resting-state fMRI data were preprocessed using the Data Processing Assistant for Resting-State fMRI (DPARSF) pipeline~\cite{yan2010dparsf}. Specifically, for each scan, the initial 10 time points were removed to allow magnetization stabilization. Subsequent preprocessing steps included slice time correction, head motion correction, and regression of confounding covariates such as white matter signal, ventricular signal, and head motion parameters. The resulting data were then normalized to the Montreal Neurological Institute (MNI) space, spatially smoothed using a Gaussian kernel with a full width at half maximum (FWHM) of 4 mm, and bandpass filtered in the frequency range of 0.01–0.1 Hz. 
Finally, mean rs-fMRI time series (\ie,  blood-oxygen-level dependent signals) were extracted from 116 regions-of-interest (ROIs) per subject, which were defined by the Automated Anatomical Labeling (AAL) atlas. 
Based on these regional BOLD signals, we construct an FC network for each subject.

\textbf{DTI Preprocessing}. 
All DTI data were preprocessed using the PANDA toolbox ~\cite{cui2013panda} following a standardized workflow. Preprocessing steps included: extracting brain tissue to remove non-brain tissue; correcting the bias field to mitigate intensity inhomogeneities; correcting for eddy currents and head motion to remove distortion; and adjusting the diffusion gradient direction. High-resolution anatomical images of each subject were registered to diffusion space to ensure spatial alignment. Subsequently, brain tissue was segmented into 116 ROIs based on AAL atlas. 
For each subject, three complementary structural connectivity (SC) metrics were calculated: 
Fiber number (FN): quantifies the number of reconstructed white matter fibers between each pair of ROIs, reflecting the strength of the connection;
Fractional anisotropy (FA): captures the directional consistency of water diffusion within white matter tracts, indicating the integrity of the microstructure; 
Fiber length (FL): measures the average length of streamlines between connected ROIs, providing geometric information about brain pathways. 
These metrics are concatenated to represent each ROI. 
\fi 

\if false 
\textbf{ASL Preprocessing}. 
All ASL data were processed using the ASLtbx toolbox~\cite{wang2008empirical} following a standardized workflow. 
Preprocessing steps include: 
extracting brain tissue to remove non-brain voxels; 
correcting the bias field to mitigate intensity inhomogeneities; 
performing ASL-specific motion correction with a zig-zag de-trending algorithm to remove both head motion and label–control alternation artifacts; 
co-registering the mean ASL image to each subject's high-resolution T1-weighted anatomical scan and applying the resulting transforms to all time points; 
spatially smoothing the realigned volumes with a Gaussian kernel (FWHM of 4 mm) to suppress high-frequency noise; 
and generating a brain mask from the mean control image to exclude residual non-brain signal. 
Quantitative cerebral blood flow (CBF) maps (mL/100g/min) are then calculated via voxel-wise control–label subtraction 
and post-labeling delay from the specific device (labeling duration = 1.8 s; labeling efficiency = 0.85; 3T), with M0 
calibration based on segmented white matter and CSF images. 
Subsequently, each subject's mean CBF map is normalized to the Montreal Neurological Institute (MNI) space for inter-subject comparability and parcellated into 116 ROIs according to the AAL atlas. 
Finally, radiomics features were extracted from these ROIs using the PyRadiomics toolbox~\cite{van2017computational}, producing regional-level features for downstream analysis.  
\fi


\subsection{Experimental Setup}


\subsubsection{Competing Method}

\if false
Three prediction tasks will be performed, including 
(1) \textbf{Task 1}: ANI vs. CN classification on HAND with fMRI and DTI data from 68 ANI subjects and 69 CN subjects; 
(2) \textbf{Task 2}: SMC vs. CN classification on ADNI with fMRI and DTI data from 46 SMC subjects and 48 CN subjects; and 
(3) \textbf{Task 3}: SMC vs. CN classification on ADNI with fMRI and ASL data from 29 SMC subjects and 15 CN subjects.
\fi 

\if false
Conventional machine learning methods, parameter setting;
Conventional GNN methods, parameter setting;
Hyerbolic GNN methods, parameter setting;
\fi

We will compare our HKGF$_1$ (with HKGCN as backbone) and HKGF$_2$ (with HKGAT as backbone) against three classical machine learning approaches, including support vector machine (\textbf{SVM})~\cite{pisner2020support}, \textbf{XGBoost}~\cite{chen2016xgboost}, and random forest (\textbf{RF})~\cite{breiman2001random} with concatenated SC and FC graph features as input. 
We will further compare our methods with eight state-of-the-art deep learning methods that automatically extract brain SC and FC features: 
graph convolutional network (\textbf{GCN})~\cite{kipf2017gcn},   
graph isomorphism network (\textbf{GIN})~\cite{xu2018gin},
graph attention network (\textbf{GAT})~\cite{velivckovic2017graph}, 
\textbf{Transformer}~\citep{vaswani2017attention}, 
graph sample and aggregate (\textbf{GraphSAGE})~\cite{hamilton2017inductive},      
\textbf{BrainNetCNN}~\cite{kawahara2017brainnetcnn},
\textbf{BrainGNN}~\cite{li2021braingnn}, and 
hyperbolic graph convolutional neural network (\textbf{HGCN})~\cite{chami2019hgcn}.

\begin{table*}[!tbp]
\setlength{\abovecaptionskip}{0pt}
\setlength{\belowcaptionskip}{0pt}
\setlength{\abovedisplayskip}{0pt}
\setlength{\belowdisplayskip}{0pt}
    \centering
        \caption{Results (\%) of 
        different methods in ANI vs. CN classification on HAND with fMRI and DTI data. `$^*$' denotes that a competing method and HKGF$_1$ are significantly different ($p<0.05$ via $t$-test), while `$^\dag$' denotes that a competing method and HKGF$_2$ are significantly different. 
}
\renewcommand{\arraystretch}{0.5}
    \setlength\tabcolsep{1.5pt}{\scalebox{1}{
    \begin{tabular}{l|ccccccc|c|c}
        \toprule

          Method & AUC & ACC & F1 & BAC& SEN & SPE & PRE &$p$-value &$p$-value \\
\midrule 
          SVM~\cite{pisner2020support} & $51.96\pm{3.25}$ &$50.08\pm{3.01}$   & $48.70\pm{3.99}$  & $51.57\pm{3.55}$  & $47.50\pm{4.61}$  & $55.63\pm{4.05}$ & $52.58\pm{4.40}$ & $<0.001^*$ & $<0.001^\dag$ \\
         RF~\cite{breiman2001random}& $54.77\pm{3.05}$ & $49.25\pm{3.05}$ & $48.49\pm{3.49}$ & $50.63\pm{3.66}$ & $49.75\pm{5.80}$ & $51.49\pm{9.83}$ & $50.88\pm{3.82}$  & $<0.001^*$ & $<0.001^\dag$ \\
       XGBoost~\cite{chen2016xgboost} & $57.49\pm{6.34}$ & $54.61\pm{5.21}$ & $54.53\pm{5.76}$ & $54.97\pm{5.01}$ & $55.70\pm{7.13}$ & $54.23\pm{5.06}$ & $55.62\pm{5.54}$  &  $<0.001^*$ &  $<0.001^\dag$\\
         \cmidrule{1-10}                  
       GCN-EF~\cite{kipf2017gcn} & $65.24\pm{5.59}$ & $64.83\pm{3.36}$  & $61.97\pm{6.88}$ & $65.54\pm{4.08}$ & $64.74\pm{10.14}$ & $66.34\pm{10.34}$  & $69.09\pm{5.05}$  & {$0.039^*$}  & $0.028^\dag$ \\   
       GCN-LF~\cite{kipf2017gcn} & $61.06\pm{2.71}$ & $63.16\pm{1.46}$  & $62.88\pm{2.66}$ & $63.35\pm{1.45}$ & $67.44\pm{6.54}$ & $59.27\pm{7.35}$  & $65.52\pm{4.04}$  & $<0.001^*$ & {$0.011^\dag$} \\       
      GAT-EF~\cite{velivckovic2017graph} & $65.10\pm{5.17}$ & $64.79\pm{3.07}$ & $61.91\pm{7.13}$  & $64.38\pm{3.19}$ & $65.60\pm{9.61}$ & $63.15\pm{6.56}$ & $67.63\pm{3.08}$  & {$0.019^*$}  & $0.005^\dag$\\
       GAT-LF~\cite{velivckovic2017graph} & $62.61\pm{3.78}$ & $64.23\pm{3.49}$ & $66.48\pm{5.90}$  & $60.77\pm{5.75}$ & 
       $64.21\pm{2.42}$ &$67.94\pm{11.26}$ &  $67.58\pm{4.33}$  & $0.046^*$ & $<0.001^\dag$\\

    Transformer-EF~\cite{vaswani2017attention} & $54.92\pm{4.36}$ & $59.13\pm{2.69}$ & $57.97\pm{5.95}$ & $59.46\pm{1.52}$ & $64.56\pm{9.28}$ & $54.36\pm{10.01}$ & $57.90\pm{8.06}$  & $<0.001^*$ & $<0.001^\dag$ \\
     Transformer-LF~\cite{vaswani2017attention} & $55.49\pm{3.88}$ & $60.56\pm{3.81}$ & $53.86\pm{2.31}$ & $59.54\pm{2.00}$ & $64.81\pm{9.42}$ & $64.26\pm{12.69}$ &  $63.29\pm{5.59}$ & $<0.001^*$ & $<0.001^\dag$ \\
        
         GraphSAGE-EF~\cite{hamilton2017inductive} & $63.11\pm3.63$ & $64.23\pm{3.98}$ & $64.24\pm{6.10}$ & $64.21\pm{4.23}$ & $71.11\pm{10.18}$ & $57.32\pm{4.89}$ & $64.04\pm{3.10}$  & {$0.025^*$}  & $0.032^\dag$\\
        GraphSAGE-LF~\cite{hamilton2017inductive} & $62.31\pm{2.30}$ & $64.23\pm{1.90}$  & $62.70\pm{2.58}$ & $63.60\pm{1.46}$ & $64.01\pm{5.54}$ & $63.18\pm{5.55}$ & $64.21\pm{1.85}$  & $0.048^*$  &  $0.024^\dag$\\

          GIN-EF~\cite{xu2018gin}  & $62.90\pm{5.34}$ & $65.12\pm{4.34}$  & $62.51\pm{8.48}$ & $64.54\pm{4.62}$ & $66.07\pm{10.43}$ & $63.00\pm{9.74}$ & $68.73\pm{9.00}$  & $0.022^*$  & {$0.033^\dag$}\\

          GIN-LF~\cite{xu2018gin}  & $ 61.30\pm{2.42}$ & $63.63\pm{2.80}$  & $61.05\pm{3.60}$ & $63.74\pm{1.94}$ & $64.91\pm{7.98}$ & $62.56\pm{11.56}$ & $69.01\pm{8.68}$  & $0.038^*$  & $<0.001^\dag$ \\
   
       BrainNetCNN-EF~\cite{kawahara2017brainnetcnn}  & $54.73\pm{4.46}$ & 
       $58.25\pm{1.83}$ & $57.97\pm{4.93}$ & $58.20\pm{1.96}$ & $63.12\pm{13.76}$ & $53.29\pm{16.81}$& $59.33\pm{4.92}$ & $<0.001^*$ 
       & $<0.001^\dag$\\
    BrainNetCNN-LF~\cite{kawahara2017brainnetcnn}  & $54.98\pm{5.88}$ & 
       $62.05\pm{4.76}$ & $57.79\pm{7.47}$ & $62.01\pm{4.94}$ & $61.08\pm{13.80}$ & $62.95\pm{13.16}$& $63.76\pm{11.95}$ & $<0.001^*$ & $<0.001^\dag$\\
            {BrainGNN-EF}~\cite{li2021braingnn} & $56.11\pm{7.60}$ & $62.08\pm{5.03}$ & $60.83\pm{3.86}$ & $62.73\pm{4.21}$ & $65.15\pm{8.33}$ & $60.31\pm{12.85}$ & $65.39\pm{5.38}$  & $0.013^*$ &  $0.048^\dag$ \\
            \textcolor{black}{BrainGNN-LF}~\cite{li2021braingnn} & $58.83\pm{5.08}$ & $62.58\pm{3.64}$ & $54.31\pm{8.33}$ & $61.39\pm{4.10}$ & $55.16\pm{13.76}$ & $67.62\pm{12.85}$ & $60.16\pm{6.79}$  & $0.009^*$ & $<0.001^{\dag}$\\
        HGCN-EF~\cite{chami2019hgcn}  & $64.08\pm{5.42}$ & $63.66\pm{3.95}$ & $65.18\pm{7.17}$ & $63.81\pm{3.81}$ & $72.35\pm{10.23}$ & $55.27\pm{4.29}$ & $64.64\pm2.18$  &  $<0.001^*$ & $<0.001^\dag$ \\ 
        HGCN-LF~\cite{chami2019hgcn}  & $60.68\pm{1.58}$ & $65.12\pm{1.15}$ & $61.02\pm{4.13}$ & $65.04\pm{1.26}$ & $59.52\pm{8.43}$ & $70.57\pm{7.10}$ & $69.74\pm{5.21}$  & $<0.001^*$  & $<0.001^\dag$\\ 
          \cmidrule{1-10}
   HKGF$_1$~(Ours) & ${68.83}\pm{2.55}$ & ${71.53}\pm{2.59}$ & $\bm{69.72}\pm{5.15}$ &${71.54}\pm{2.53}$ & $\bm{71.30}\pm{8.73}$ & ${71.78}\pm{4.18}$ & ${73.62}\pm{1.72}$ & --  &-- \\
        HKGF$_2$ (Ours)& $\bm{69.74}\pm{1.77}$ & $\bm{71.86}\pm{2.39}$ & ${68.73}\pm{2.54}$ &$\bm{72.36}\pm{1.40}$ &${70.45}\pm{7.57}$ & $\bm{74.27}\pm{7.64}$ & $\bm{76.45}\pm{4.99}$ &--   & --\\
        \bottomrule
    \end{tabular}}}
    \label{tab_ANIfMRIDTI}
\end{table*}

\if false
\begin{table*}[!tbp]
\setlength{\abovecaptionskip}{0pt}
\setlength{\belowcaptionskip}{0pt}
\setlength{\abovedisplayskip}{0pt}
\setlength{\belowdisplayskip}{0pt}
    \centering
        \caption{Results (\%) of 
        different methods in SMC vs. CN classification on ADNI with fMRI and DTI data. 
        `$^*$' denotes that a competing method and HKGF$_1$ are significantly different ($p<0.05$ via $t$-test), while `$^\dag$' denotes that a competing method and HKGF$_2$ are significantly different.{\color{red}UPDATE!}
}
\renewcommand{\arraystretch}{0.5}
    \setlength\tabcolsep{1.5pt}{\scalebox{1}{
    \begin{tabular}{l|ccccccc|c|c}
        \toprule

          Method & AUC & ACC & F1 & BAC& SEN & SPE & PRE &$p$-value$^*$ &$p$-value$^\dag$  \\
\midrule 
          SVM~\cite{pisner2020support} & $55.08\pm{5.90}$ & $57.63\pm{2.60}$ & $59.92\pm{2.05}$ & $58.51\pm{1.25}$ & $59.19\pm{3.97}$ & $57.83\pm{5.54}$ & $64.42\pm{2.09}$  & $<0.001^*$ &  $<0.001^\dag$\\
         RF~\cite{breiman2001random} & $67.75\pm{3.54}$ & $60.42\pm{2.99}$ & $67.29\pm{2.42}$ & $60.80\pm{2.58}$ & $76.52\pm{3.68}$ & $45.08\pm{7.38}$ & $63.28\pm{2.39}$  & $<0.001^*$ & $<0.001^\dag$\\
       XGBoost~\cite{chen2016xgboost} & $70.54\pm{4.87}$ & $63.53\pm{3.87}$ & $67.20\pm{4.43}$ & $64.26\pm{4.74}$ & $70.42\pm{3.50}$&$58.10\pm{7.11}$ & $67.49\pm{4.75}$   &  $<0.001^*$  &  $<0.001^\dag$ \\
         \cmidrule{1-10}                   
    GCN-EF~\cite{kipf2017gcn} & $74.72\pm{4.31}$ & $73.65\pm{4.98}$ & $73.65\pm{6.22}$ & $75.41\pm{3.37}$ & $71.89\pm{10.27}$ & $78.93\pm{10.15}$ & $\bm{83.19}\pm{5.95}$  &  $<0.001^*$  &  $<0.001^\dag$\\
    {GCN-LF}~\cite{kipf2017gcn}&$70.02\pm{2.08}$& $72.22\pm{1.92}$ &$74.92\pm{2.49}$&$71.48\pm{1.78}$&$78.33\pm{4.59}$&$64.63\pm{5.20}$& $74.73\pm{2.46}$  & $<0.001^*$
  & $<0.001^\dag$\\

    GAT-EF~\cite{velivckovic2017graph} & $73.87\pm{1.47}$  & $75.84\pm{2.34}$  & $76.06\pm{3.16}$ & $75.62\pm{1.78}$ & $74.44\pm{7.55}$ & $76.81\pm{7.15}$ &  $80.86\pm{2.61}$  & $<0.001^*$ &  $<0.001^\dag$\\
    GAT-LF~\cite{velivckovic2017graph}&$77.67\pm{2.80}$&$75.68\pm{2.98}$& $77.48\pm{3.24}$ &$75.43\pm{2.60}$& $78.53\pm{6.14}$ & $72.32\pm{5.15}$ & $79.10\pm{1.55}$  & $0.330$  & $<0.001^\dag$\\  

    Transformer-EF~\cite{vaswani2017attention}&$66.95\pm{2.26}$&$68.64\pm{3.96}$&${67.83}\pm{2.53}$&$69.36\pm{4.09}$&$65.80\pm{6.51}$&$72.92\pm{13.80}$& $79.61\pm{9.27}$  & $0.001^*$   & $<0.001^\dag$\\
    
    {Transformer-LF}~\cite{vaswani2017attention}& $70.89\pm{7.81}$ &$71.66\pm{8.10}$&$75.26\pm{6.85}$&$71.02\pm{7.85}$&$81.76\pm{10.43}$&$60.29\pm{17.73}$&$74.99\pm{8.62}$  & $<0.001^*$  & $0.048^\dag$ \\
    
    GraphSAGE-EF~\cite{hamilton2017inductive}  & $73.82\pm{3.16}$ & $69.05\pm{1.62}$ & $69.56\pm{2.48}$ & $70.93\pm{3.10}$ & $67.67\pm{5.18}$ & $74.19\pm{8.02}$ & $77.38\pm{4.51}$  & $<0.001^*$ & $<0.001^\dag$ \\
    GraphSAGE-LF~\cite{hamilton2017inductive}  & $75.70\pm{3.77}$ & $72.55\pm{2.49}$ & $72.62\pm{4.00}$ & $74.34\pm{2.51}$ & $69.37\pm{6.21}$ & $\bm{79.32}\pm{3.82}$ & $81.70\pm{3.04}$  & $0.012^*$ & $<0.001^\dag$ \\   

    GIN-EF~\cite{xu2018gin} &$74.17\pm{4.10}$&$70.96\pm{3.14}$&$71.37\pm{3.43}$&$72.86\pm{2.46}$&$68.38\pm{4.31}$&$77.33\pm{3.44}$&$79.27\pm{3.22}$ & $<0.001^*$  & $<0.001^\dag$\\      
       
    GIN-LF~\cite{xu2018gin}  & $70.32\pm{2.34}$ & $69.42\pm{1.41}$ & $67.10\pm{3.04}$ & $69.88\pm{1.66}$ & $63.39\pm{6.81}$ & $76.37\pm{8.36}$ & $79.55\pm{3.86}$  & $<0.001^*$ & $<0.001^\dag$ \\ 
        
    {BrainNetCNN-EF}~\cite{kawahara2017brainnetcnn}  & $68.72\pm{8.60}$ & $61.85\pm{3.43}$ & $57.85\pm{5.58}$ & $62.92\pm{4.22}$ & $57.15\pm{9.20}$ & $68.70\pm{15.94}$ & $66.55\pm{11.12}$  & $<0.001^*$  &  $<0.001^\dag$ \\
    BrainNetCNN-LF~\cite{kawahara2017brainnetcnn} & $71.95\pm{6.53}$ & $74.07\pm{4.55}$& $75.25\pm{6.36}$&$75.74\pm{5.01}$&$75.66\pm{12.75}$&$75.82\pm{8.53}$& $80.80\pm{2.79}$  &  $<0.001^*$ & $<0.001^\dag$ \\   
    
    BrainGNN-EF~\cite{li2021braingnn} &&&&&&&   & \\ 
    BrainGNN-LF~\cite{li2021braingnn} &&&&&&&   & \\ 
    
    HGCN-EF~\cite{chami2019hgcn}  & $71.10\pm{4.04}$ & $69.28\pm{2.52}$ & $68.58\pm{4.29}$ & $69.79\pm{2.39}$ & $65.00\pm{3.12}$ & $74.58\pm{3.28}$ & $77.70\pm{2.51}$  & $<0.001^*$  & $<0.001^\dag$\\    
    HGCN-LF~\cite{chami2019hgcn} &$69.81\pm{5.59}$&$69.58\pm{3.90}$&$70.37\pm{3.58}$&$69.91\pm{4.43}$&$67.33\pm{5.22}$&$72.49\pm{11.04}$&$78.24\pm{7.50}$  & $<0.001^*$ & $<0.001^\dag$\\
    
    \cmidrule{1-10}
    HKGF$_1$~(Ours)   & ${75.72}\pm{2.31}$ & ${77.19}\pm{2.39}$ & ${79.11}\pm{3.16}$ &${76.80}\pm{0.89}$ &${82.04}\pm{5.35}$ & ${71.55}\pm{5.27}$ & ${78.75}\pm{0.58}$ & --  &-- \\
    HKGF$_2$~(Ours) & $\bm{80.42}\pm{2.25}$ & $\bm{81.26}\pm{2.62}$ & $\bm{82.65}\pm{2.64}$ &$\bm{80.58}\pm{2.11}$ &$\bm{84.81}\pm{4.08}$ & ${76.35}\pm{3.97}$ & ${81.75}\pm{3.11}$ &  -- & --\\
        \bottomrule
    \end{tabular}}}
    \label{tab_SMCfMRIDTI}
\end{table*}
\fi

\if false 
{\color{red}
(1) \textbf{SVM}: 
Multiple node-level features (\ie, degree centrality, clustering coefficient, betweenness centrality, and eigenvector centrality) 
and graph-level features (dimension: $2,336$) 
are extracted from the SC and FC graphs from each subject. 
This feature vector is then used as input to a linear SVM  (with the default parameter $C=1$) for classification. 
}

(2) \textbf{XGBoost}: Similar to SVM, 
the same set of node-level and graph-level features extracted from brain SC and FC graphs are concatenated into a feature vector for representing each subject. The vectorized representation is then classified using XGBoost with default parameters. 

(3) \textbf{Random Forest}: This method uses the same vectorized features as SVM and XGBoost, followed by a random forest classifier (using default settings). 

\if false
{\color{red}
(4) \textbf{LSTM}: 
}
(5) \textbf{STGCN}: 
STGCN is designed to concurrently extract spatial and temporal features from fMRI time series using spatiotemporal graph convolution units (GCUs). Initially, the fMRI time series of each subject is input into two sequential GCNs to encode spatial dependencies across brain regions. The spatial embeddings are subsequently processed by temporal convolutional units, implemented via standard 1D convolutional layers, to model temporal dynamics across successive fMRI volumes. Two GCUs are stacked to comprehensively capture joint spatiotemporal patterns, and a graph-level representation is obtained through a pooling operation, followed by two fully connected layers for prediction.
\fi

(4) \textbf{GCN}: 
It is designed specifically for handling graph-structured data. 
With a brain FC/SC graph as input, two stacked graph convolutional layers are applied to update node features, and a readout operation is used to derive a graph-level representation. This is followed by a two-layer multilayer perceptron (MLP) with two fully-connected layers and a Softmax function) for classification.

(5) \textbf{GIN}: 
This method processes the input FC/SC graph using two GIN layers that leverage the Weisfeiler-Lehman graph isomorphism test for feature learning. A graph-level feature vector is obtained via a readout operation, followed by an MLP for classification. 
Note that the node and edge features of GIN are similar to those in GCN. 

(6) \textbf{GAT}: 
GAT incorporates an attention mechanism to learn adaptive edge weights between ROIs. We use two graph attention layers with $4$ attention heads each to capture spatial features. A readout operation generates a graph-level vector, followed by an MLP for prediction.

\begin{table*}[!tbp]
\setlength{\abovecaptionskip}{0pt}
\setlength{\belowcaptionskip}{0pt}
\setlength{\abovedisplayskip}{0pt}
\setlength{\belowdisplayskip}{0pt}
    \centering
        \caption{Results (\%) of 
        different methods in SMC vs. CN classification on the ADNI dataset with resting-state fMRI and DTI. 
        The term marked `$^*$' denotes that HKGF$_1$ and a competing method are significantly different ($p<0.05$ via $t$-test), while `$^\dag$' denotes that HKGF$_2$ and a competing method are significantly different. {\color{red}UPDATE!}
}
\renewcommand{\arraystretch}{0.5}
    \setlength\tabcolsep{1.5pt}{\scalebox{1}{
    \begin{tabular}{l|ccccccc|c|c}
        \toprule

          Method & AUC & ACC & F1 & BAC& SEN & SPE & PRE &$p$-value$^*$ &$p$-value$^\dag$  \\
\midrule 
          SVM~\cite{pisner2020support} & $55.08\pm{5.90}$ & $57.63\pm{2.60}$ & $59.92.48\pm{2.05}$ & $58.51\pm{1.25}$ & $59.19\pm{3.97}$ & $57.83\pm{5.54}$ & $64.42\pm{2.09}$  &  & \\
         RF~\cite{breiman2001random} & $67.75\pm{3.54}$ & $60.42\pm{2.99}$ & $67.29\pm{2.42}$ & $60.80\pm{2.58}$ & $76.52\pm{3.68}$ & $45.08\pm{7.38}$ & $63.28\pm{2.39}$  &  & \\
       XGBoost~\cite{chen2016xgboost} & $70.54\pm{4.87}$ & $63.53\pm{3.87}$ & $67.20\pm{4.43}$ & $64.26\pm{4.74}$ & $70.42\pm{3.50}$&$58.10\pm{7.11}$ & $67.49\pm{4.75}$   &  & \\
         \cmidrule{1-10}                   
    GCN-EF~\cite{kipf2017gcn} & $74.72\pm{4.31}$ & $73.65\pm{4.98}$ & $73.65\pm{6.22}$ & $75.41\pm{3.37}$ & $71.89\pm{10.27}$ & $78.93\pm{10.15}$ & $83.19\pm{5.95}$  &   & \\
    \textcolor{red}{GCN-LF}~\cite{kipf2017gcn}&$68.61\pm{2.35}$& $66.30\pm{2.76}$ &$64.42\pm{8.30}$&$67.01\pm{2.02}$&$61.67\pm{13.18}$&$72.36\pm{12.09}$& $77.91\pm{8.54}$  & 
  & \\

    GAT-EF~\cite{velivckovic2017graph} & $73.87\pm{1.47}$  & $75.84\pm{2.34}$  & $76.06\pm{3.16}$ & $75.62\pm{1.78}$ & $74.44\pm{7.55}$ & $76.81\pm{7.15}$ &  $80.86\pm{2.61}$  &  & \\
    GAT-LF~\cite{velivckovic2017graph}&$77.67\pm{2.80}$&$75.68\pm{2.98}$& $77.48\pm{3.24}$ &$75.43\pm{2.60}$& $78.53\pm{6.14}$ & $72.32\pm{5.15}$ & $79.10\pm{1.55}$  &   & \\  

    Transformer-EF~\cite{vaswani2017attention}&$76.94\pm{1.34}$&$77.06\pm{1.21}$&$\bm{78.19}\pm{2.14}$&$77.06\pm{2.21}$&$77.85\pm{6.72}$&$76.27\pm{8.98}$& $81.63\pm{2.38}$  &   & \\
    
    Transformer-LF~\cite{vaswani2017attention}& $69.49\pm{6.48}$ &$63.98\pm{4.66}$&$57.42\pm{10.05}$&$64.03\pm{5.38}$&$65.58\pm{13.52}$&$62.48\pm{16.83}$&$55.23\pm{12.02}$  &   & \\

    GraphSAGE-EF~\cite{hamilton2017inductive}  & $76.83\pm{1.35}$ & $72.76\pm{2.17}$ & $73.78\pm{2.75}$ & $72.36\pm{4.63}$ & $72.56\pm{5.95}$ & $72.16\pm{4.63}$ & $78.87\pm{3.51}$  &  & \\
    GraphSAGE-LF~\cite{hamilton2017inductive}  & $73.91\pm{3.41}$ & $70.48\pm{3.31}$ & $68.70\pm{7.77}$ & $71.74\pm{2.33}$ & $65.18\pm{11.90}$ & $73.80\pm{9.48}$ & $80.24\pm{4.66}$  &  & \\

    GIN-EF~\cite{xu2018gin}  & $76.94\pm{2.04}$ & $78.00\pm{1.90}$ & $79.56\pm{2.06}$ & $79.56\pm{1.46}$ & $81.09\pm{2.70}$ & $78.02\pm{3.40}$ & $81.94\pm{1.24}$  &  & \\ 
    
    GIN-LF~\cite{xu2018gin} &$71.28\pm{11.22}$&$69.02\pm{8.45}$&$66.72\pm{12.28}$&$69.57\pm{8.75}$&$64.01\pm{13.80}$&$75.12\pm{6.80}$&$75.05\pm{9.81}$ &   & \\       
    
    BrainNetCNN-EF~\cite{kawahara2017brainnetcnn}  & $63.44\pm{5.95}$ & $67.04\pm{2.63}$ & $69.48\pm{6.28}$ & $66.35\pm{2.87}$ & $72.93\pm{12.36}$ & $59.78\pm{12.74}$ & $71.30\pm{3.94}$  &  & \\
    BrainNetCNN-LF~\cite{kawahara2017brainnetcnn} & $72.31\pm{6.93}$ & $69.86\pm{5.19}$& $68.30\pm{6.55}$&$69.51\pm{5.11}$&$65.97\pm{7.71}$&$73.06\pm{7.82}$& $78.60\pm{5.19}$  & \\   
    
    BrainGNN-EF~\cite{li2021braingnn} &&&&&&&   & \\ 
    BrainGNN-LF~\cite{li2021braingnn} &&&&&&&   & \\ 
    
    HGCN-EF~\cite{chami2019hgcn}  & $71.11\pm{3.55}$ & $68.29\pm{1.55}$ & $66.39\pm{6.53}$ & $68.80\pm{1.38}$ & $64.67\pm{13.09}$ & $72.93\pm{14.34}$ & $79.80\pm{8.06}$  &  & \\    
    HGCN-LF~\cite{chami2019hgcn} &$66.36\pm{4.31}$&$66.06\pm{1.83}$&$65.52\pm{3.03}$&$66.39\pm{2.43}$&$63.67\pm{6.71}$&$69.11\pm{9.84}$&$75.83\pm{5.38}$  &  & \\
    
    \cmidrule{1-10}
    HKGF$_1$~(Ours)   & $\bm{75.72}\pm{2.31}$ & ${77.19}\pm{2.39}$ & ${79.11}\pm{3.16}$ &${76.80}\pm{0.89}$ &${82.04}\pm{5.35}$ & $\bm{71.55}\pm{5.27}$ & $\bm{78.75}\pm{0.58}$ & --  & \\
    HKGF$_2$~(Ours) & $\bm{80.17}\pm{1.87}$ & $\bm{80.68}\pm{0.81}$ & ${81.84}\pm{1.84}$ &$\bm{80.66}\pm{1.08}$ &$\bm{83.03}\pm{5.36}$ & ${78.29}\pm{4.14}$ & ${82.45}\pm{2.55}$ & --  & \\
        \bottomrule
    \end{tabular}}}
    \label{tab_SMCfMRIDTI}
\end{table*}

(7) \textbf{Transformer}: It is a deep learning model based on self-attention techniques for processing sequential data.
In this method, each FC/SC network is regarded as the query, key, and value matrices,  
which are fed into the Transformer. 
We then leverage self-attention to dynamically weigh the importance of different regions in the network, capturing dependencies among brain regions. The learned graph representations are then fed into an MLP for prediction.

(8) \textbf{GraphSAGE}: It is tailored for learning on graph-structured data. When applied to brain networks, it generates node embeddings by sampling and aggregating information from each node's local neighborhood. Similar to GCN, the node and edge features in GraphSAGE are defined based on functional/structural connectivity: the feature vector for the $j$-th ROI corresponds to the 
$j$-th row of the connectivity matrix, capturing its connections to other regions, while edge features reflect the strength of inter-regional functional/structural connectivity. For feature extraction, two GraphSAGE layers are employed, followed by a pooling operation to derive a graph-level representation, followed by using a two-layer MLP for prediction.

(9) \textbf{BrainNetCNN}: 
The input of BrainNetCNN is a brain FC or SC graph representing the functional/structural connectome. 
Its architecture comprises three specialized convolutional filters (\ie, edge-to-edge, edge-to-node, and node-to-graph) which are tailored to capture the hierarchical topological structure of brain networks from edge-level interactions to global graph-level representations. This is followed by a two-layer MLP for classification.

(10) \textbf{BrainGNN}: 
It includes an node-selection layer to highlight salient nodes. Using the constructed SC/FC graph as input, it employs two node-aware graph convolutional layers to learn the node representations followed by node pooling. The resulting node embeddings are aggregated via a readout operation to form a graph-level representation, which is then classified using a two-layer MLP. 

(11) \textbf{HGCN}: Similar to our HKGCN and HKGAT, this method extends GCNs to non-Euclidean hyperbolic space to model hierarchical and tree-like structures common in real-world graphs. Specifically, it first embeds an input SC/FC graph into hyperbolic space. 
Node features, representing regional brain activity patterns, are then mapped to this non-Euclidean space using an initial projection. 
Two hyperbolic graph convolutional layers are then applied to learn hierarchical representations of brain regions by aggregating information across the graph using Möbius-based operations. Finally, a hyperbolic-to-Euclidean projection followed by a pooling operation is used to generate a graph-level embedding for MLP-based classification. 
\fi 

\if false
{\color{red}
The two methods (\ie, LSTM and STGCN) take BOLD signals derived from fMRI as input, since they use a series of dynamic graphs (generated through a sliding window strategy with a window size of $40$~\cite{hindriks2016can}) as input. 
}
\fi

The three machine learning models (SVM, RF, and XGBoost) use the same input features (dimension: 2,336) per subject, including node-level (\eg, degree centrality, clustering coefficient, betweenness, eigenvector centrality) and graph-level features from each subject's SC and FC graphs. 
The deep learning methods use the same multimodal graph inputs as ours and a two-layer MLP for prediction, with 32 neurons in the first layer and 2 neurons in the second. 
We employ the default settings for these competing methods and diligently ensure their training hyperparameters are aligned with ours to ensure a fair comparison.

The competing deep learning methods incorporate both early fusion (EF) and late fusion (LF) strategies for multimodal feature integration. 
Taking the GCN as a representative example, early fusion (denoted as \textbf{GCN-EF}) begins by concatenating the node-level FC and SC features to form a fused node representation. 
Simultaneously, the adjacency matrices of the FC and SC graphs are summed 
to construct a single fused adjacency matrix. 
This composite graph (consisting of the fused node feature matrix and the fused adjacency matrix) is then fed into a GCN for feature extraction, followed by a two-layer MLP for final prediction. 
In contrast, the late fusion strategy (denoted as \textbf{GCN-LF}) processes the SC and FC graphs separately using two independent GCN models to learn modality-specific features. The resulting features are then concatenated and passed through the same two-layer MLP to generate the final output. 
\if false
Late Fusion VS. Early Fusion -- 
Late Fusion: Use two parallel GCNs for DTI and fMRI graph feature learning, and then concatenate these two features for MLP-based prediction; 
Early Fusion: Use one GCN for feature learning and MLP for prediction, with concatenated SC and FC feature as node feature and summed and normalized adjacency matrix as input adjacency. 
\fi


\subsubsection{Prediction Task and Evaluation Metric}

Twp prediction tasks will be performed on two target cohorts (\ie, ADNI and HAND), including 
(1) \textbf{Task 1}: SMC vs. CN classification on ADNI with fMRI and ASL data from 29 SMC subjects and 15 CN subjects, and 
(2) \textbf{Task 2}: ANI vs. CN classification on HAND with fMRI and DTI data from 68 ANI subjects and 69 CN subjects.  
To address the small data issue, 
We employ a \emph{transfer learning strategy} by first pretraining a deep learning model on large-scale source data and then fine-tuning it on the target data to perform downstream target tasks. 
Using self-supervised contrastive learning ~\cite{fang2025action}, we pretrain the deep models (\ie, GCN, GIN, GAT, Transformer, GraphSAGE, BrainNetCNN, BrainGNN, HGCN, HKGCN, and HKGAT) on the auxiliary source data containing 3,806 fMRI scans from 
ABIDE~\cite{di2014autism}, REST-meta-MDD Consortium~\cite{yan2019reduced}, and ADHD-200~\cite{adhd2012adhd}. 
We then fine-tune these pretrained models on target data for prediction. 
There is no data overlap between the auxiliary and target data.  
All the pretrained models can be accessed oneline\footnote{\url{https://github.com/mxliu/ACTION-Software}}. 
\if false
For our HKGCN and HKGAT, we initialize the parameters of their graph convolutional layers using those from those pretrained from GCN and GAT models. 
This initialization facilitates more stable optimization and enhances the model's generalization capability.
{\color{purple}New Solution: Qianqian will help pretrain the three models (\ie, HGCN, HKGCN, and HKGAT) on large data. 
}
\fi


All methods are evaluated using a standard 5-fold cross-validation (CV) strategy on the two target datasets. 
To reduce variability due to random data splits, the CV process is repeated five times with different random seeds. 
The average results and standard deviation across the five runs are reported. 
Performance evaluation is conducted using seven metrics: area under the ROC curve (AUC), accuracy (ACC), F1 score (F1), balanced accuracy (BAC), sensitivity (SEN), specificity (SPE), and precision (PRE).

\subsection{Results and Analysis}
\subsubsection{Task 1: SMC vs. CN Prediction on ADNI}
We present the results of all methods for SMC vs. CN classification on the ADNI dataset using fMRI and DTI data in Table~\ref{tab_SMCfMRIDTI}. 
We conduct pairwise $t$-test on the results achieved by HKGF$_1$ and each competing method to assess their difference significance, with significant results ($p$$<$$0.05$) marked as `$^*$' in Table~\ref{tab_SMCfMRIDTI}, while those for the comparison between HKGF$_2$ and each competing methods are marked as `$^\dag$'. 
Several important insights can be derived from Table~\ref{tab_SMCfMRIDTI}. 
\emph{First}, both versions of our method (HKGF$_1$ and HKGF$_2$) generally outperform competing approaches, with HKGF$_2$ achieving the highest overall performance (AUC: 80.42\%, ACC: 81.26\%). 
The slight superiority of HKGF$_2$ over HKGF$_1$ may be attributed to the attention mechanism employed in the HKGAT backbone, which enables it to capture informative dependencies that the HKGCN backbone in HKGF$_1$ lacks. 
\emph{Second}, HKGF$_1$ and HKGF$_2$ yield statistically significant improvements over traditional methods such as SVM and XGBoost, and various GCN-based models across most evaluation metrics. 
\emph{Third}, compared to transformer-based models (\eg, Transformer-EF) and GraphSAGE variants, our methods achieve higher sensitivity and specificity, indicating more balanced classification performance. 
\emph{Additionally}, $t$-test results confirm that HKGF$_1$ and HKGF$_2$ significantly outperform most competing methods, further validating the effectiveness of our HKGF framework in SMC detection. 


\subsubsection{Task 2: ANI vs. CN Prediction on HAND}

Table~\ref{tab_ANIfMRIDTI} reports the results of all methods in ANI vs. CN classification on HAND. 
We draw several key observations from this table. 
\emph{First}, deep learning approaches generally outperform traditional machine learning methods (\ie, SVM, RF, and XGBoost), highlighting the advantages of data-driven graph feature extraction.
\emph{Second}, our HKGF$_1$ and HKGF$_2$ achieve the best performance (\eg, AUC of 69.74\% and ACC of 71.85\% by HKGF$_2$), indicating their effectiveness in modeling cross-modality local-to-global interactions and hierarchical dependencies among ROIs. 
\emph{Third}, our methods significantly outperform state-of-the-art models such as BrainNetCNN and BrainGNN. 
This improvement may be attributed to the SC-FC coupling module in HKGF, as opposed to simple SC and FC feature concatenation used in the two competing methods.  
\emph{Finally}, the overall performance across all methods in this task is generally lower than that reported for SMC identification in Table~\ref{tab_SMCfMRIDTI}. 
This difference may stem from the nature of the cohorts: while SMC subjects experience subjective cognitive complaints, ANI patients often do not exhibit noticeable cognitive symptoms, making ANI identification inherently more challenging. 

\if false
\subsubsection{Task 1: ANI vs. CN Prediction on HAND}

Table~\ref{tab_ANIfMRIDTI} reports the results of 21 methods in ANI vs. CN classification on HAND. 
We conduct pairwise $t$-test on the results achieved by HKGF$_1$ and each competing method to assess their difference significance, with significant results ($p$$<$$0.05$) marked as `$^*$' in Table~\ref{tab_ANIfMRIDTI}, while those for the comparison between HKGF$_2$ and each competing methods are marked as `$^\dag$'.   
We draw several key observations from Table~\ref{tab_ANIfMRIDTI}. 
\emph{First}, deep learning approaches generally outperform traditional machine learning methods (\ie, SVM, RF, and XGBoost), highlighting the advantages of data-driven SC and FC graph feature extraction.
\emph{Second}, our proposed HKGF$_1$ and HKGF$_2$ models achieve the best  performance (\eg, AUC of 68.83\% and ACC of 71.53\% by HKGF$_1$), indicating their effectiveness in modeling cross-modality local-to-global interactions and hierarchical dependencies among ROIs. 
\emph{Third}, our methods significantly outperform state-of-the-art models such as BrainNetCNN and BrainGNN. 
This improvement may be attributed to the SC-FC coupling module in HKGF, as opposed to simple SC and FC feature concatenation used in the two competing methods.  
\emph{Finally}, $t$-test results confirm that our HKGF$_1$ and HKGF$_2$ significantly outperform most competing methods, further validating the effectiveness of the proposed HKGF framework in identifying ANI patients within the HAND cohort.

\subsubsection{Task 2: SMC vs. CN Prediction on ADNI}
We present the results of all methods for SMC vs. CN classification on the ADNI dataset using fMRI and DTI data in Table~\ref{tab_SMCfMRIDTI}.
\emph{On one hand}, the performance across methods is generally higher than that reported for ANI identification in Table~\ref{tab_ANIfMRIDTI}.
This may be attributed to the fact that SMC subjects exhibit subjective cognitive decline, while ANI patients typically lack noticeable cognitive symptoms at this stage, making ANI identification more challenging.
\emph{On the other hand}, both versions of our method (HKGF$_1$ and HKGF$_2$) consistently outperform competing approaches, with HKGF$_1$ achieving the highest overall performance (\eg, AUC: 80.17\%, ACC: 80.68\%). 
The slight superiority of HKGF$_2$ over HKGF$_1$ may be attributed to the attention mechanism employed in the HKGAT backbone, which enables it to capture informative dependencies that the HKGCN backbone in HKGF$_1$ lacks. 
\emph{Furthermore}, HKGF$_1$ and HKGF$_2$ yield statistically significant improvements ($p<0.05$) over traditional methods such as SVM and XGBoost, as well as various GCN-based models across most evaluation metrics. 
\emph{Additionally}, compared to transformer-based models (\eg, Transformer-EF) and GraphSAGE variants, our methods achieve higher sensitivity and specificity, indicating more balanced classification performance. 
In this task, HKGF$_2$ achieves the best results among all models, closely followed by HKGF$_1$, with significant gains in accuracy and balanced performance (BAC, SEN, SPE) over both conventional and GNN/transformer approaches. 
These findings further validate the superiority of our HKGF framework.
\fi

\begin{figure*}[!t]
\setlength{\abovecaptionskip}{0pt}
\setlength{\belowcaptionskip}{0pt}
\setlength\abovedisplayskip{0pt}
\setlength\belowdisplayskip{0pt}
\centering
\includegraphics[width=1\textwidth]{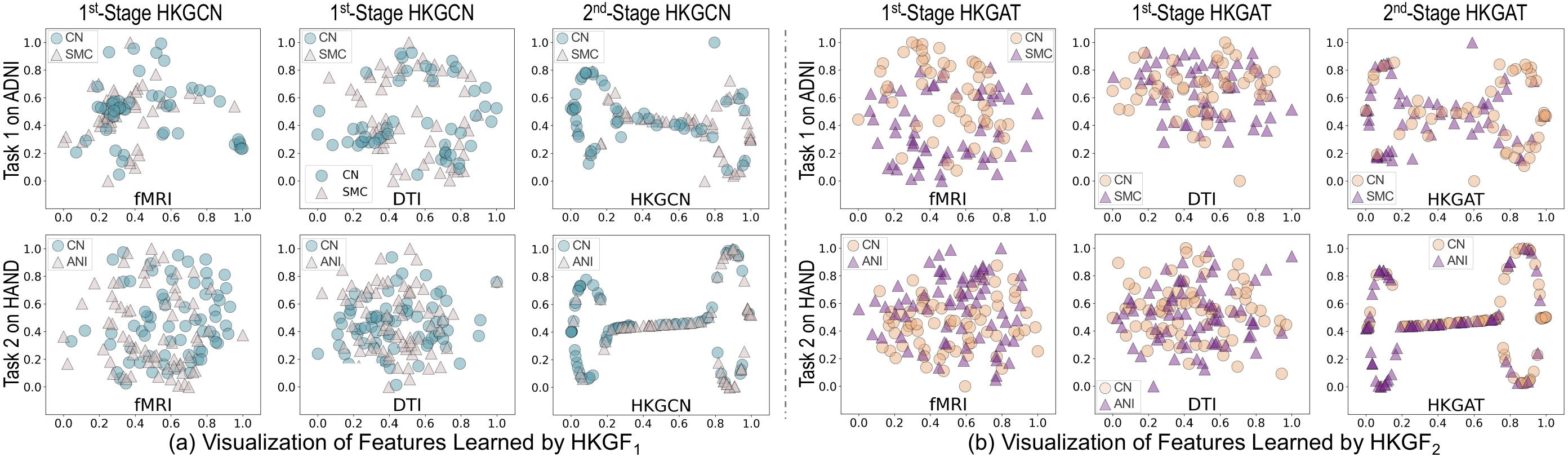}
\caption{The t-SNE~\cite{van2008visualizing} visualization of features output by (a) HKGCN backbone and (b) HKGAT backbone in single-modality graph representation learning (1st stage) and cross-modality coupling (2nd stage) modules of the proposed HKGF$_1$ and HKGF$_2$, respectively.  
Shown for (top) Task 1: SMC vs. CN classification on the ADNI dataset and (bottom) ANI vs. CN classification on the HAND dataset using fMRI and DTI data.} 
\label{fig_tSNE}
\end{figure*}

\subsection{Visualization of Learned Graph Representation}
For the two prediction tasks, we visualize the graph features learned by the first-stage and second-stage HKGCN backbones of HKGF$_1$ using t-SNE~\cite{van2008visualizing} in Fig.~\ref{fig_tSNE}~(a), where the three columns correspond to FC graphs derived from fMRI, SC graphs from DTI, and the SC-FC coupling graphs produced by our cross-modality coupling module.  
Figure~\ref{fig_tSNE}~(b) presents a comparison of the feature distributions learned by the first-stage and second-stage HKGAT backbones of the proposed HKGF$_2$. 
These visualizations show that the feature representations of FC and SC graphs show considerable overlap, while the fused SC-FC coupling graph features can more clearly distinguish between categories. 
It is important to recognize that the two classification tasks are inherently challenging due to the subtlety and complexity of asymptomatic neurocognitive disorders in HIV patients and the subjective nature of SMCs. 
This complexity leads to limited feature separability between categories.

\subsection{Identified Discriminative Brain Regions}

\begin{figure}[!tbp]
\setlength{\abovecaptionskip}{-2pt}
\setlength{\belowcaptionskip}{0pt}
\setlength{\abovedisplayskip}{0pt}
\setlength{\belowdisplayskip}{0pt}
\centering
\includegraphics[width=0.495\textwidth]{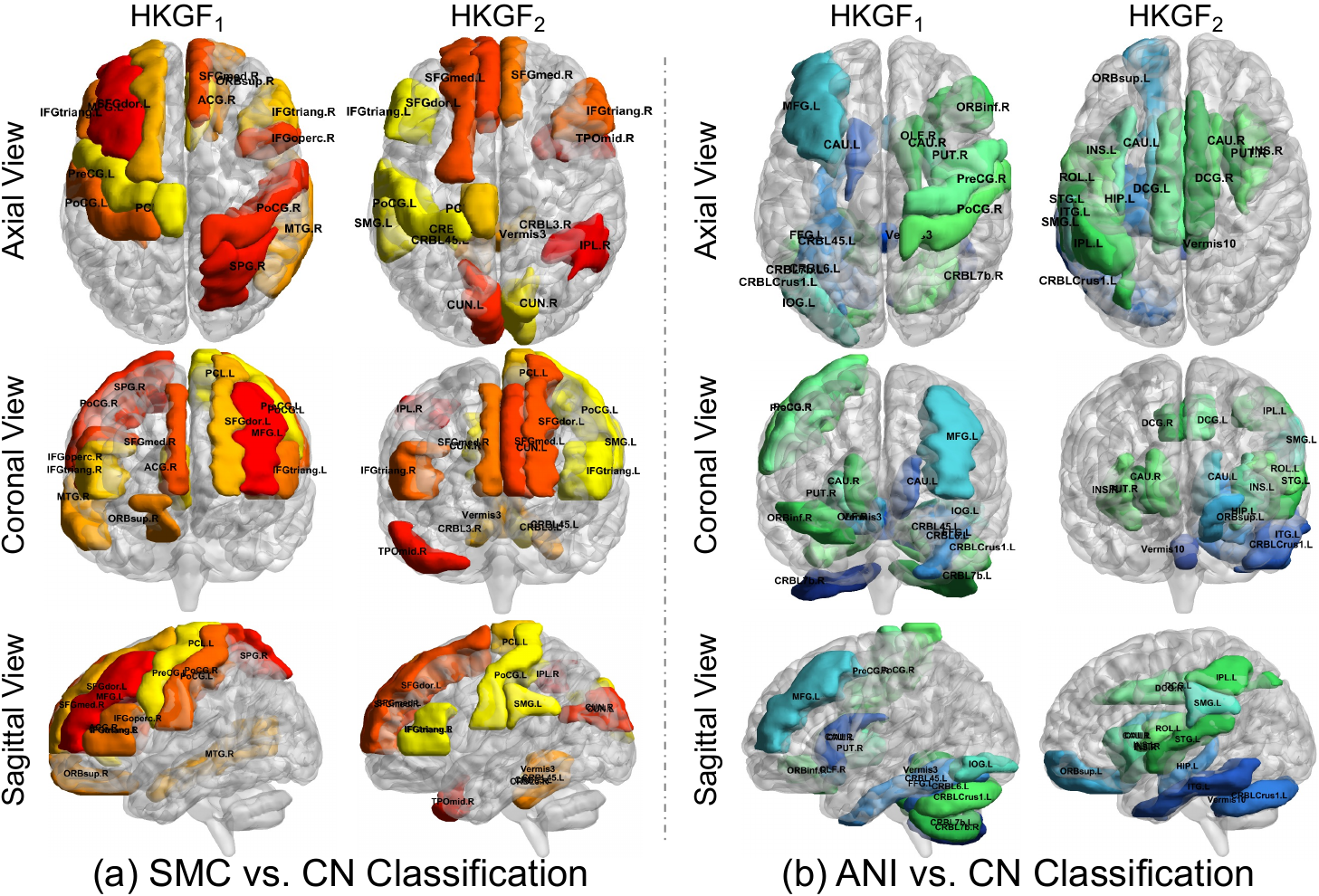} 
\caption{Discriminative brain regions identified by HKGF$_1$ (with HKGCN as backbone) and HKGF$_2$ (with HKGAT as backbone) in the tasks of  (a) SMC vs. CN classification and (b) ANI vs. CN classification with fMRI and DTI data. Different colors represent distinct regions. 
} 
\label{fig_showSFC}
\end{figure}

We further study the  discriminative brain regions identified by our methods in the two tasks based on SC-FC coupling graph embeddings 
through the  
BrainNet Viewer~\cite{xia2013brainnet}. 
The visualization results are shown in Fig.~\ref{fig_showSFC}. 
The SC-FC coupling features are derived from our SC-FC coupling modules in HKGF$_1$ and HKGF$_2$. 
Specifically, for HKGF$_1$ with the HKGCN backbone (without any attention techniques), we compute Pearson's correlation coefficients between paired ROIs of these embeddings 
and use $t$-test (significance level: 0.05) to identify significant connectivity differences between the positive (\eg, SMC) and negative (\ie, CN) groups across five folds. 
We select the top 15 discriminative connectivities, involving 14 ROIs in SMC vs. CN classification and 16 ROIs in ANI vs. CN classification.  
For HKGF$_2$ with the HKGAT backbone, we rely on the attention matrix learned by the second-stage HKGAT to obtain the 
discriminative ROIs.

Figure~\ref{fig_showSFC}~(a) indicates that, despite some variability, both methods consistently identify six overlapping ROIs in the SMC vs. CN classification, including SFGdor.L, IFGtriang.L, IFGtriang.R, SFGmed.R, PoCG.L, PCL.L. 
These regions are primarily located in the frontal and parietal lobes, which are known to be involved in executive function, attention, and early memory processing. 
Their involvement aligns with previous studies reporting frontal-parietal dysfunction in individuals with subjective memory concerns~\cite{rodda2011subjective,woolgar2010fluid,jamali2019dorsolateral}, suggesting these ROIs may serve as early neural markers of cognitive decline. 
Figure~\ref{fig_showSFC}~(b) shows that, in the ANI vs. CN classification, both methods consistently identify four important ROIs (CAU.L, CAU.R, PUT.R, and CRBLCrus1.L). 
These findings align with previous studies~\cite{janssen2017resting,wang2018altered,wright2016putamen}, further supporting the reliability of our methods in distinguishing ANI from CNs. 
Notably, connectivity between the left cerebellar Crus I (CRBLCrus1.L) and the left inferior temporal gyrus (ITG.L) suggests involvement of cerebellar–cortical circuits in semantic processing. 
Given the cerebellum's role in motor-cognitive integration and ITG's function in language and semantics, this connection may reflect disruptions in integrative processing in HAND~\cite{wang2018altered}. 
We also found strong coupling between the left hippocampus (HIP.L) and ITG.L, a circuit commonly linked to memory function and frequently implicated in HAND pathology. 
This supports prior evidence that memory-related circuits are among the most vulnerable in HAND~\cite{lopez1997biochemical}. 
Additionally, the observed interaction between the left superior orbital frontal gyrus (ORBsup.L) and Vermis 10 may relate to emotional and cognitive regulation, as both regions are involved in affective and executive processing. 
Finally, the connection between the left caudate nucleus (CAU.L) and Vermis 10 points to dysfunction in frontal–subcortical loops. 
The caudate is involved in attention control, while the cerebellar vermis supports behavioral coordination. Disruption in this pathway may underlie deficits in attention and motor function commonly seen in HAND, emphasizing the involvement of cortico-subcortical networks in the disorder~\cite{wang2018altered}.

\section{Discussion}
\label{S5}

\begin{table*}[!tbp]
\setlength{\abovecaptionskip}{0pt}
\setlength{\belowcaptionskip}{0pt}
\setlength{\abovedisplayskip}{0pt}
\setlength{\belowdisplayskip}{0pt}
    \centering
        \caption{Performance (\%) of the proposed methods (\ie, HKGF$_1$ and HKGF$_2$) and their ablated variants in two prediction tasks.}
\renewcommand{\arraystretch}{1}
\setlength\tabcolsep{0.3pt}
{\scalebox{0.91}{
    \begin{tabular}{l|ccccccc |ccccccc}
\toprule
\multirow{2}{*}{Method} & \multicolumn{7}{c|}{SMC vs. CN Classification on ADNI}  & \multicolumn{7}{c}{ANI vs. CN Classification on HAND}              
\\
\cline{2-15}
& AUC & ACC & F1 & BAC& SEN & SPE & PRE  & AUC & ACC & F1 & BAC& SEN & SPE & PRE \\
\midrule
{\color{black}HKGF-G}  & $74.27_{2.23}$& $74.69_{3.93}$& $75.99_{2.87}$ & $75.50_{3.05}$ & $77.30_{4.08}$&$73.70_{6.46}$ & $79.85_{3.43}$ 
& $63.23_{3.98}$ & $67.72_{3.98}$ & $65.10_{8.62}$ & $67.84_{3.12}$ & $64.31_{13.86}$ & $71.38_{8.19}$ & $72.83_{3.02}$ \\

{\color{black}HKGF-A} & $79.65_{2.24}$ & $77.15_{2.35}$ & $78.13_{1.79}$ & $77.14_{1.83}$ & $77.74_{3.48}$& $76.53_{5.23}$& $80.91_{3.10}$ 
& $65.73_{2.74}$& $66.44_{1.90}$ & $62.58_{4.93}$ & $66.64_{0.83}$ & $63.48_{10.24}$& $69.80_{9.71}$ & $71.81_{7.21}$ 
\\

{\color{black}{HKGF-K}} & $72.15_{5.68}$ & $74.25_{4.74}$ & $74.98_{6.02}$ & $74.41_{4.16}$ & $73.00_{9.46}$ & $75.82_{4.30}$& $81.53_{2.24}$ 
& $65.39_{2.59}$& $64.66_{2.69}$ & $63.90_{3.59}$ & $64.69_{2.52}$ & $66.11_{3.39}$ & $63.27_{1.95}$ & $65.98_{1.54}$
\\

\hline
{\color{black}HKGF$_1$w/oC} & $68.15_{12.23}$ & $67.62_{5.35}$ & $68.75_{6.30}$ & $68.06_{5.40}$ & $65.00_{10.87}$& $71.11_{12.86}$ & $74.72_{8.69}$ 
& $60.64_{4.90}$& $64.38_{2.60}$ & $64.22_{5.78}$ & $64.47_{2.62}$ & $70.33_{9.09}$ & $58.61_{5.72}$ & $64.37_{2.73}$ 
\\
{\color{black}HKGF$_1$w/oH} & $73.24_{2.87}$& $75.50_{3.69}$ & $76.22_{2.50}$ & $75.25_{3.31}$ & $77.20_{3.00}$& $73.31_{7.16}$& $78.97_{3.21}$ 
& $63.54_{4.62}$& $64.80_{3.23}$ & $61.80_{7.07}$ & $64.87_{3.15}$ & $63.04_{11.80}$ & $66.70_{8.77}$ & $68.68_{5.22}$ 
\\
{\color{black}HKGF$_2$w/oC} & $78.00_{2.01}$ & $77.71_{2.14}$ & $79.86_{2.09}$ & $78.17_{2.50}$ & $\bm{82.28}_{4.05}$ & $74.07_{4.55}$ & $80.76_{3.38}$
& $67.41_{1.48}$& $67.02_{2.58}$ & $66.97_{6.14}$ & $66.61_{2.12}$ & $73.01_{11.68}$ & $60.20_{10.59}$ & $66.68_{5.65}$
\\
{\color{black}HKGF$_2$w/oH} & $79.72_{0.90}$ & $80.35_{0.14}$& $80.06_{1.06}$ & $79.70_{0.67}$ & $77.62_{3.71}$ & $\bm{81.78}_{4.26}$& $85.56_{3.03}$
& $69.39_{1.70}$& $67.92_{1.24}$ & $65.43_{5.85}$ & $68.17_{1.06}$ & $67.51_{12.70}$ & $68.83_{10.76}$ & $71.67_{8.95}$
\\
\hline
    HKGF$_1$   & ${76.22}_{2.31}$ & ${76.30}_{1.99}$ & ${77.59}_{2.54}$ &${78.10}_{2.13}$ &${78.34}_{4.77}$ & ${77.87}_{3.15}$ & ${81.32}_{1.99}$
    & ${68.83}_{2.55}$ & ${71.53}_{2.59}$ & $\bm{69.72}_{5.15}$ &${71.54}_{2.53}$ & $\bm{71.30}_{8.73}$ & ${71.78}_{4.18}$ & ${73.62}_{1.72}$  
    \\
    HKGF$_2$ (Ours) & $\bm{80.36}_{1.33}$ & $\bm{81.83}_{1.35}$ & $\bm{82.10}_{2.21}$ &$\bm{80.92}_{1.26}$ &${81.98}_{5.92}$ & ${79.86}_{5.98}$ & $\bm{84.96}_{2.76}$ 
    & $\bm{69.74}_{1.77}$ & $\bm{71.86}_{2.39}$ & ${68.73}_{2.54}$ &$\bm{72.36}_{1.40}$ &${70.45}_{7.57}$ & $\bm{74.27}_{7.64}$ & $\bm{76.45}_{4.99}$ \\
\bottomrule
    \end{tabular}}}
\label{tab_ablation2Task}
\end{table*}

\begin{figure*}[!tbp]
\setlength{\abovecaptionskip}{0pt}
\setlength{\belowcaptionskip}{0pt}
\setlength{\abovedisplayskip}{0pt}
\setlength{\belowdisplayskip}{0pt}
\centering
\subfigure{\includegraphics[width=1\linewidth]{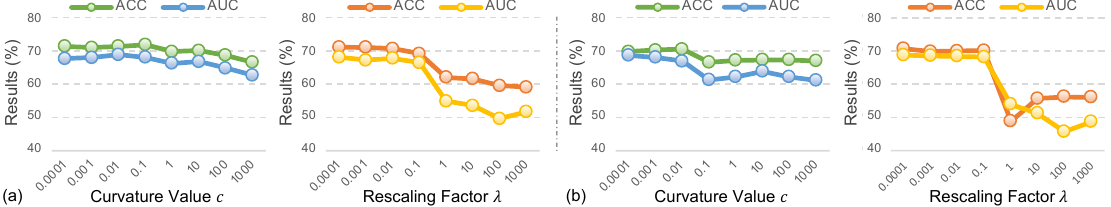}}
\caption{Influence of the two hyperparameters ($c$ and $\lambda$) on the performance of (a) HKGF$_1$ and (b) HKGF$_2$ for ANI vs. CN classification on HAND.}
\label{fig_parameter}
\end{figure*}

\subsection{Ablation Study}
To evaluate the influence of key components, we compare our two HKGF implementations  (\ie, HKGF$_1$ and HKGF$_2$) with their variants:  
(1) \textbf{HKGF-G} that uses GCN instead of HKGCN/HKGAT as backbone,  
(2) \textbf{HKGF-A} with GAT as backbone, 
(3) \textbf{HKGF-K} with HGCN as backbone,  
(4) \textbf{HKGF$_1$w/oC}, which removes SC-FC coupling and concatenates HKGCN-extracted fMRI and DTI features,  
(5) \textbf{HKGF$_2$w/oC} that removes SC-FC coupling and concatenates HKGAT-extracted fMRI and DTI features, and 
(6) \textbf{HKGF$_1$w/oH} and \textbf{HKGF$_2$w/oH} that use MLP (with 2 fully-connected
layers) instead of HNN as a predictor.

As reported in Table \ref{tab_ablation2Task}, 
HKGF$_1$ and HKGF$_2$ achieve the overall best performance compared to their variants. 
\emph{First}, the three variants (HKGF-G, HKGFC-A, and HKGF-K)  show moderate performance degradation compared to HKGF$_1$ and HKGF$_2$ across both tasks. 
This demonstrates our method’s ability to model hierarchical dependencies among ROIs, an aspect not captured by HKGF-G and HKGF-A, as well as their superior capacity to capture cross-modality local-to-global interactions compared to HKGF-K.   
\emph{Second}, the performance drop of HKGF$_1$w/oC HKGF$_2$w/oC highlights the importance of explicit SC-FC coupling in feature fusion (as we do in this work).   
\emph{Additionally}, the slight performance decline observed in HKGF$_1$w/oH and HKGF$_2$w/oH implies that our HNN used in  HKGF$_1$ and HKGF$_2$ may better aligns with the hyperbolic geometry compared to the standard MLP predictor.

\subsection{Hyperparameter Analysis}
We further investigate the influence of two key hyperparameters in HKGCN and HKGAT: 
(1) curvature value $c$ which determines the geometry of hyperbolic space, and (2) $\lambda$ which scales the cosine function.   
To assess their impact, we conduct experiments on HAND for ANI vs. CN classification.  
In Fig.~\ref{fig_parameter}, we report the AUC and ACC performance of HKGF$_1$ and HKGF$_2$ under different settings of $c$ and $\alpha$. 
These results reveal that the curvature parameter $c$ has a relatively minor effect on the two models' performance in terms of ACC and AUC. 
This suggests that our methods are robust to variations in the underlying geometric space. 
In contrast, the scaling factor $\lambda$
has a more noticeable impact, and the performance remains stable when $\lambda<1$ in both cases, indicating a reasonable range for tuning without degrading predictive performance. 

\if false
\begin{table}[!tbp]
\setlength{\abovecaptionskip}{0pt}
\setlength{\belowcaptionskip}{0pt}
\setlength{\abovedisplayskip}{0pt}
\setlength{\belowdisplayskip}{0pt}
    \centering
        \caption{Performance (\%) of the proposed methods (HKGF$_1$ and HKGF$_2$) and their ablated variants in SMC vs. CN classification on  ADNI.}
\renewcommand{\arraystretch}{1}
    \setlength\tabcolsep{0.3pt}{\scalebox{0.85}{
    \begin{tabular}{l|ccccccc}
\toprule
Method & AUC & ACC & F1 & BAC& SEN & SPE & PRE \\
\midrule
{\color{black}HKGF-G}  & $74.27_{2.23}$& $74.69_{3.93}$& $75.99_{2.87}$ & $75.50_{3.05}$ & $77.30_{4.08}$&$73.70_{6.46}$ & $79.85_{3.43}$ \\
{\color{black}HKGF-A} & & &  &  & & &  \\
{\color{red}{HKGF-K}} & $71.96_{3.35}$ & $75.01_{2.31}$ & $76.22_{3.23}$ & $75.13_{2.35}$ & $74.47_{7.40}$ & $75.60_{8.41}$& $80.78_{4.71}$ \\
\hline
{\color{black}HKGF$_1$w/oC} & $68.15_{12.23}$ & $67.62_{5.35}$ & $68.75_{6.30}$ & $68.06_{5.40}$ & $65.00_{10.87}$& $71.11_{12.86}$ & $74.72_{8.69}$  \\
{\color{black}HKGF$_1$w/oH} & $73.24_{2.87}$& $75.50_{3.69}$ & $76.22_{2.50}$ & $75.25_{3.31}$ & $77.20_{3.00}$& $73.31_{7.16}$& $78.97_{3.21}$ \\
{\color{black}HKGF$_2$w/oC} & & &  &  & & &  \\
{\color{black}HKGF$_2$w/oH} & & & &  &  & &  \\
\hline
    HKGF$_1$   & ${75.72}_{2.31}$ & ${77.19}_{2.39}$ & ${79.11}_{3.16}$ &${76.80}_{0.89}$ &${82.04}_{5.35}$ & ${71.55}_{5.27}$ & ${78.75}_{0.58}$ \\
    HKGF$_2$ & $\bm{80.42}_{2.25}$ & $\bm{81.26}_{2.62}$ & $\bm{82.65}_{2.64}$ &$\bm{80.58}_{2.11}$ &$\bm{84.81}_{4.08}$ & ${76.35}_{3.97}$ & ${81.75}_{3.11}$ \\
\bottomrule
    \end{tabular}}}
\label{tab:ablationstudy}
\end{table}

\begin{table}[!tbp]
\setlength{\abovecaptionskip}{0pt}
\setlength{\belowcaptionskip}{0pt}
\setlength{\abovedisplayskip}{0pt}
\setlength{\belowdisplayskip}{0pt}
    \centering
        \caption{Performance (\%) of the proposed methods (HKGF$_1$ and HKGF$_2$) and their ablated variants in ANI vs. CN classification on  HAND.}
\renewcommand{\arraystretch}{1}
    \setlength\tabcolsep{0.3pt}{\scalebox{0.85}{
    \begin{tabular}{l|ccccccc}
\toprule
Method & AUC & ACC & F1 & BAC& SEN & SPE & PRE \\
\midrule
{\color{black}HKGF-G}  & $63.23_{3.98}$ & $67.72_{3.98}$ & $65.10_{8.62}$ & $67.84_{3.12}$ & $64.31_{13.86}$ & $71.38_{8.19}$ & $72.83_{3.02}$ \\
{\color{black}HKGF-A} & $65.73_{2.74}$& $66.44_{1.90}$ & $62.58_{4.93}$ & $66.64_{0.83}$ & $63.48_{10.24}$& $69.80_{9.71}$ & $71.81_{7.21}$\\ 
{\color{black}{HKGF-K}} & $65.39_{2.59}$& $64.66_{2.69}$ & $63.90_{3.59}$ & $64.69_{2.52}$ & $66.11_{3.39}$ & $63.27_{1.95}$ & $65.98_{1.54}$\\ 
\hline
{\color{black}HKGF$_1$w/oC} & $60.64_{4.90}$& $64.38_{2.60}$ & $64.22_{5.78}$ & $64.47_{2.62}$ & $70.33_{9.09}$ & $58.61_{5.72}$ & $64.37_{2.73}$\\ 
{\color{black}HKGF$_1$w/oH} & $63.54_{4.62}$& $64.80_{3.23}$ & $61.80_{7.07}$ & $64.87_{3.15}$ & $63.04_{11.80}$ & $66.70_{8.77}$ & $68.68_{5.22}$\\ 
{\color{black}HKGF$_2$w/oC} & $67.41_{1.48}$& $67.02_{2.58}$ & $66.97_{6.14}$ & $66.61_{2.12}$ & $73.01_{11.68}$ & $60.20_{10.59}$ & $66.68_{5.65}$\\ 
{\color{black}HKGF$_2$w/oH} & $68.30_{1.62}$& $69.22_{0.81}$ & $67.26_{2.40}$ & $70.27_{1.66}$ & $70.42_{7.61}$ & $70.11_{5.57}$ & $72.27_{1.93}$\\ 
\hline
 HKGF$_1$ & ${68.83}_{2.55}$ & ${71.53}_{2.59}$ & $\bm{69.72}_{5.15}$ &${71.54}_{2.53}$ & $\bm{71.30}_{8.73}$ & ${71.78}_{4.18}$ & ${73.62}_{1.72}$  \\
    HKGF$_2$& $\bm{69.74}_{1.77}$ & $\bm{71.86}_{2.39}$ & ${68.73}_{2.54}$ &$\bm{72.36}_{1.40}$ &${70.45}_{7.57}$ & $\bm{74.27}_{7.64}$ & $\bm{76.45}_{4.99}$ \\
\bottomrule
    \end{tabular}}}
\label{tab:ablationstudy}
\end{table}
\fi

\if false
\begin{table*}[!tbp]
\setlength{\abovecaptionskip}{0pt}
\setlength{\belowcaptionskip}{0pt}
\setlength{\abovedisplayskip}{0pt}
\setlength{\belowdisplayskip}{0pt}
    \centering
        \caption{Results (\%) of 
        different methods in SMC vs. CN classification on ADNI  {\color{blue}with fMRI and ASL data}. 
       `$^*$' denotes that a competing method and HKGF$_1$ are significantly different ($p<0.05$ via $t$-test), while `$^\dag$' denotes that a competing method and HKGF$_2$ are significantly different.{\color{red}UPDATE!}
}
\renewcommand{\arraystretch}{0.5}
    \setlength\tabcolsep{2pt}{\scalebox{1}{
    \begin{tabular}{l|ccccccc|c|c}
        \toprule

          Method & AUC & ACC & F1 & BAC& SEN & SPE & PRE &$p$-value$^*$ &$p$-value$^{\dag}$ \\
\midrule 
    SVM~\cite{pisner2020support}&$64.59\pm{10.60}$&$66.50\pm{9.17}$&$69.10\pm{13.67}$&$62.89\pm{9.22}$&$70.10\pm{14.98}$&$55.67\pm{4.01}$&$75.63\pm{10.84}$& & \\
    RF~\cite{breiman2001random}&$74.96\pm{4.14}$&$75.61\pm{4.90}$&$79.68\pm{4.32}$&$75.05\pm{6.15}$&$77.77\pm{5.24}$&$72.33\pm{11.34}$&$85.11\pm{6.02}$& & \\ 
    XGBoost~\cite{chen2016xgboost}&$64.61\pm{5.19}$&$64.67\pm{3.08}$&$65.48\pm{6.48}$ &$66.69\pm{2.77}$&$60.70\pm{7.38}$&$72.67\pm{6.93}$&$80.38\pm{4.14}$&& \\
    
    \cmidrule{1-10}                   
    GCN-EF~\cite{kipf2017gcn}&$70.18\pm{9.06}$&$71.33\pm{7.41}$&$74.71\pm{5.55}$&$72.51\pm{4.36}$&$75.69\pm{4.93}$&$69.33\pm{8.38}$&$80.68\pm{5.08}$& & \\  
    GCN-LF~\cite{kipf2017gcn}&$62.11\pm{4.17}$&$65.56\pm{4.02}$&$70.33\pm{4.45}$&$63.80\pm{5.22}$&$68.93\pm{7.55}$&$58.67\pm{12.82}$&$80.59\pm{5.64}$& & \\ 
    
    GAT-EF~\cite{velivckovic2017graph}&$74.68\pm{6.14}$&$77.50\pm{2.68}$&$78.97\pm{3.38}$& $\bm{81.44}\pm{2.87}$&$70.22\pm{4.83}$&$92.67\pm{3.46}$&$95.58\pm{1.72}$& \\
    GAT-LF~\cite{velivckovic2017graph}& $75.14\pm{4.76}$& $78.83\pm{4.04}$ &$80.72\pm{2.97}$&$80.40\pm{2.98}$&$73.46\pm{6.41}$&$87.33\pm{9.47}$&$94.32\pm{3.41}$& \\

    Transformer-EF~\cite{vaswani2017attention}&$73.10\pm{7.98}$&$76.17\pm{4.04}$&$78.14\pm{3.95}$&$78.20\pm{4.16}$&$72.39\pm{4.34}$&$84.00\pm{8.63}$&$91.72\pm{2.95}$& \\
    Transformer-LF~\cite{vaswani2017attention}&$72.78\pm{7.41}$&$74.39\pm{8.84}$&$75.79\pm{10.45}$&$76.81\pm{9.27}$&$69.63\pm{14.35}$&$84.00\pm{13.62}$&$93.11\pm{4.31}$& \\
        
    GraphSAGE-EF~\cite{hamilton2017inductive}&$73.21\pm{4.80}$&$76.06\pm{3.48}$&$80.80\pm{3.34}$&$74.21\pm{5,25}$&$78.43\pm{6.35}$&$70.00\pm{13.54}$&$86.84\pm{4.27}$& \\
    GraphSAGE-LF~\cite{hamilton2017inductive}&$66.82\pm{8.01}$&$70.78\pm{7.42}$&$72.41\pm{12.10}$&$70.34\pm{5.91}$&$73.69\pm{17.75}$&$67.00\pm{16.30}$&$76.18\pm{7.71}$& \\

    GIN-EF~\cite{xu2018gin}&$67.16\pm{5.02}$&$75.44\pm{6.34}$&$76.74\pm{9.35}$&$76.86\pm{3.37}$&$72.40\pm{11.39}$&$81.33\pm{11.14}$&$87.46\pm{11.02}$ & \\
    GIN-LF~\cite{xu2018gin}&$68.43\pm{8.79}$&$74.28\pm{7.98}$&$72.35\pm{12.13}$&$78.27\pm{5.48}$&$66.67\pm{15.41}$&$89.87\pm{7.08}$&$90.99\pm{6.55}$ & \\
    
    BrainNetCNN-EF~\cite{kawahara2017brainnetcnn}&$61.82\pm{12.12}$&$66.61\pm{5.71}$&$68.17\pm{8.40}$&$66.70\pm{3.33}$&$65.08\pm{10.91}$&$68.33\pm{8.90}$&$78.37\pm{10.02}$& \\
    BrainNetCNN-LF~\cite{kawahara2017brainnetcnn} &$73.35\pm{6.85}$&$69.56\pm{3.48}$&$70.96\pm{4.52}$&$72.07\pm{3.61}$&$62.47\pm{4.81}$&$81.67\pm{5.27}$&$90.36\pm{2.78}$ & \\   
    
    BrainGNN-EF~\cite{li2021braingnn} &&&&&&& & \\ 
    BrainGNN-LF~\cite{li2021braingnn} &&&&&&& & \\ 
    
    HGCN-EF~\cite{chami2019hgcn}&$65.29\pm{9.27}$&$63.61\pm{4.20}$&$63.63\pm{8.03}$&$67.27\pm{3.14}$&$55.87\pm{8.29}$&$78.67\pm{5.58}$&$84.51\pm{8.24}$ & & \\    
    HGCN-LF~\cite{chami2019hgcn} &$67.69\pm{4.38}$&$66.39\pm{4.12}$&$68.82\pm{6.56}$&$68.93\pm{2.90}$&$60.53\pm{9.72}$&$77.33\pm{10.01}$&$86.77\pm{5.84}$ & & \\
    
    \cmidrule{1-10}
    HKGF$_1$~(Ours)&$75.96\pm{2.37}$&$76.78\pm{3.44}$&$78.70\pm{4.82}$&$80.50\pm{1.07}$&$74.66\pm{10.01}$&$86.33\pm{9.38}$&$90.43\pm{6.04}$ & -- &--\\
    HKGF$_2$~(Ours)& $\bm{76.85}\pm{3.78}$ & $\bm{79.22}\pm{4.27}$ & ${81.12}\pm{5.32}$ & ${79.69}\pm{4.36}$ &${75.04}\pm{8.77}$ &$\bm{84.33}\pm{1.49}$  & ${93.08}\pm{1.11}$ & -- &-- \\
        \bottomrule
    \end{tabular}}}
    \label{tab_SMCfMRIASL}
\end{table*}
\fi

\if false 
\begin{table*}[!tbp]
\setlength{\abovecaptionskip}{0pt}
\setlength{\belowcaptionskip}{0pt}
\setlength{\abovedisplayskip}{0pt}
\setlength{\belowdisplayskip}{0pt}
    \centering
        \caption{Results (\%) of 
        different methods in SMC vs. CN classification on the ADNI dataset {\color{blue}with ASL and DTI}. 
        `*' denotes that the HKGF and a competing method are significantly different ($p<0.05$ via $t$-test). {\color{red}UPDATE!}
}
\renewcommand{\arraystretch}{0.8}
    \setlength\tabcolsep{4pt}{\scalebox{1}{
    \begin{tabular}{l|ccccccc|c}
        \toprule

          Method & AUC & ACC & F1 & BAC& SEN & SPE & PRE &$p$-value  \\
\midrule 
    SVM~\cite{pisner2020support}&&&&&&&\\
    RF~\cite{breiman2001random}&&&&&&&\\
    XGBoost~\cite{chen2016xgboost}&&&&&&&\\
    
    \cmidrule{1-9}                   
    GCN-EF~\cite{kipf2017gcn}&&&&&&&\\  
    GCN-LF~\cite{kipf2017gcn}&&&&&&&\\ 
    
    GAT-EF~\cite{velivckovic2017graph}&&&&&&&\\
    GAT-LF~\cite{velivckovic2017graph}&&&&&&&\\

    Transformer-EF~\cite{vaswani2017attention}&&&&&&&\\
    Transformer-LF~\cite{vaswani2017attention}&&&&&&&\\
        
    GraphSAGE-EF~\cite{hamilton2017inductive}&$71.00\pm{4.88}$&$72.22\pm{6.29}$&$75.43\pm{6.03}$&$73.56\pm{5.98}$&$75.46\pm{8.98}$&$71.67\pm{11.61}$&$84.69\pm{4.82}$\\
    GraphSAGE-LF~\cite{hamilton2017inductive}&$\bm{76.80}\pm{3.60}$&$82.73\pm{2.72}$&$83.93\pm{3.03}$&$82.45\pm{2.07}$&$83.57\pm{8.86}$&$81.33\pm{12.21}$&$86.83\pm{4.44}$\\

    GIN-EF~\cite{xu2018gin}&&&&&&&\\
    GIN-LF~\cite{xu2018gin}&&&&&&&\\
    
    BrainNetCNN-EF~\cite{kawahara2017brainnetcnn}&&&&&&&\\
    BrainNetCNN-LF~\cite{kawahara2017brainnetcnn} &&&&&&&\\   
    
    BrainGNN-EF~\cite{li2021braingnn} &&&&&&&\\ 
    BrainGNN-LF~\cite{li2021braingnn} &&&&&&&\\ 
    
    HGCN-EF~\cite{chami2019hgcn}&&&&&&& \\    
    HGCN-LF~\cite{chami2019hgcn} &&&&&&&\\
    
    \cmidrule{1-9}
    HKGCN~(Ours) &$72.53\pm{3.17}$&$80.39\pm{1.31}$&$82.71\pm{1.22}$&$82.17\pm{2.80}$&$80.35\pm{6.26}$&$84.00\pm{11.22}$&$89.91\pm{6.65}$\\
    HKGAT~(Ours) & ${74.12}\pm{2.62}$ & ${78.17}\pm{2.46}$ & ${80.37}\pm{3.05}$ &$\bm{80.40}\pm{2.55}$ &$\bm{77.80}\pm{4.53}$ & ${83.00}\pm{7.01}$ & ${88.39}\pm{3.21}$ & -- \\
        \bottomrule
    \end{tabular}}}
    \label{tab_SMCASLDTI}
\end{table*}
\fi

\subsection{Influence of Transfer Learning}
To evaluate the effectiveness of our transfer learning strategy for model pretraining, we compare the proposed models (\ie, HKGF$_1$ and HKGF$_2$) with their respective variants (called \textbf{HKGF$_1$w/oP} and \textbf{HKGF$_2$w/oP}, respectively). 
Both HKGF$_1$w/oP and HKGF$_2$w/oP are trained from scratch on the target data without any pretraining. 
The results of these methods in ANI vs. CN classification on HAND and SMC vs. CN classification on ADNI are reported in Table~\ref{tab_transfer}. 
As presented in Table~\ref{tab_transfer}, both HKGF$_1$ and HKGF$_2$ outperform their corresponding variants across most evaluation metrics. 
These variants, which are trained from scratch without pretraining, show notably lower performance. 
For instance, HKGF$_1$ improves the AUC by more than 5\% compared to HKGF$_1$w/oP on HAND, highlighting the effectiveness of our transfer learning approach.  
These results suggest that leveraging large-scale auxiliary data (as we do in this work) significantly enhances model generalization, especially when the amount of target data is limited. 
\if false
t-Test: significance
1) XX  HKGCN vs. HKGCNw/oP
2) XX HKGAT vs. HKGAT/oP
AUC, ACC, F1
\fi 

\begin{table*}[!tbp]
\setlength{\abovecaptionskip}{0pt}
\setlength{\belowcaptionskip}{0pt}
\setlength{\abovedisplayskip}{0pt}
\setlength{\belowdisplayskip}{0pt}
    \centering
        \caption{Results(\%) of the proposed methods (\ie, HKGF$_1$ and HKGF$_2$) and their variants without model pretraining in two prediction tasks.}
\renewcommand{\arraystretch}{1}
\setlength\tabcolsep{0.3pt}
{\scalebox{0.91}{
    \begin{tabular}{l|ccccccc |ccccccc}
\toprule
\multirow{2}{*}{Method} & \multicolumn{7}{c|}{SMC vs. CN Classification on ADNI}  & \multicolumn{7}{c}{ANI vs. CN Classification on HAND}            
\\
\cline{2-15}
& AUC & ACC & F1 & BAC& SEN & SPE & PRE  & AUC & ACC & F1 & BAC& SEN & SPE & PRE \\ 
\midrule

{\color{black}HKGF$_1$w/oP} & $73.75_{2.68}$&$76.29_{2.14}$ & $76.69_{5.15}$ & $75.62_{1.87}$ & $77.77_{12.56}$& ${73.48}_{11.59}$& ${79.08}_{4.87}$
&  $63.62_{3.48}$ & $68.03_{1.91}$ & $68.42_{4.13}$& $68.18_{1.85}$ & $\bm{72.81}_{8.47}$ & $63.54_{6.49}$ & $68.51_{1.54}$ \\

 HKGF$_1$~(Ours)& $\bm{76.22}_{2.31}$ & $\bm{76.30}_{1.99}$ & $\bm{77.59}_{2.54}$ &$\bm{78.10}_{2.13}$ &$\bm{78.34}_{4.77}$ & $\bm{77.87}_{3.15}$ & $\bm{81.32}_{1.99}$ 
 & $\bm{68.83}_{2.55}$ & $\bm{71.53}_{2.59}$ & $\bm{69.72}_{5.15}$ &$\bm{71.54}_{2.53}$ & ${71.30}_{8.73}$ & $\bm{71.78}_{4.18}$ & $\bm{73.62}_{1.72}$  \\
\midrule 
\textcolor{black}{HKGF$_2$w/oP } & $76.22_{1.29}$ & $77.94_{1.71}$ & $78.53_{0.67}$ & $77.45_{1.11}$ & $78.30_{1.39}$ & $76.60_{2.36}$ & $82.59_{3.30}$ 
& $66.42_{2.52}$& $65.70_{2.79}$ & $66.69_{2.76}$ &  $66.12_{2.76}$& $\bm{72.46}_{3.54}$ & $59.79_{5.06}$  & $65.57_{2.15}$ 
\\ 

        HKGF$_2$ (Ours) & $\bm{80.36}_{1.33}$ & $\bm{81.83}_{1.35}$ & $\bm{82.10}_{2.21}$ &$\bm{80.92}_{1.26}$ &$\bm{81.98}_{5.92}$ & $\bm{79.86}_{5.98}$ & $\bm{84.96}_{2.76}$ 
    & $\bm{69.74}_{1.77}$ & $\bm{71.86}_{2.39}$ & $\bm{68.73}_{2.54}$ &$\bm{72.36}_{1.40}$ &${70.45}_{7.57}$ & $\bm{74.27}_{7.64}$ & $\bm{76.45}_{4.99}$ \\
\bottomrule
\end{tabular}}}
\label{tab_transfer}
\end{table*}

\begin{table}[!tbp]
\setlength{\abovecaptionskip}{0pt}
\setlength{\belowcaptionskip}{0pt}
\setlength{\abovedisplayskip}{0pt}
\setlength{\belowdisplayskip}{0pt}
    \centering
        \caption{
        Generalization evaluation results (\%) of different methods in SMC vs. CN classification on ADNI with fMRI and ASL data.}
\scriptsize
\renewcommand{\arraystretch}{0.6}
\setlength\tabcolsep{1pt}
    \begin{tabular}{l|ccccc}
        \toprule

          Method & AUC & ACC & F1 & BAC& SEN 
          \\
\midrule 
    SVM
    &$64.59_{10.60}$&$66.50_{9.17}$&$69.10_{13.67}$&$62.89_{9.22}$&$70.10_{14.98}$ 
    \\
    RF
    &$74.96_{4.14}$&$75.61_{4.90}$&$79.68_{4.32}$&$75.05_{6.15}$&$\bm{77.77}_{5.24}$
    \\ 
    XGBoost
    &$64.61_{5.19}$&$64.67_{3.08}$&$65.48_{6.48}$ &$66.69_{2.77}$&$60.70_{7.38}$
    \\
    
    \cmidrule{1-6}                   
    GCN-EF
    &$70.18_{9.06}$&$71.33_{7.41}$&$74.71_{5.55}$&$72.51_{4.36}$&$75.69_{4.93}$
    \\  
    GCN-LF
    &$62.11_{4.17}$&$65.56_{4.02}$&$70.33_{4.45}$&$63.80_{5.22}$&$68.93_{7.55}$
    \\ 
    
    GAT-EF
    &$74.68_{6.14}$&$77.50_{2.68}$&$78.97_{3.38}$& $\bm{81.44}_{2.87}$&$70.22_{4.83}$
    \\
    
    GAT-LF
    & $75.14_{4.76}$& $78.83_{4.04}$ &$80.72_{2.97}$&$80.40_{2.98}$&$73.46_{6.41}$
    \\

    Transformer-EF
    &$73.10_{7.98}$&$76.17_{4.04}$&$78.14_{3.95}$&$78.20_{4.16}$&$72.39_{4.34}$
    \\
    Transformer-LF
    &$72.78_{7.41}$&$74.39_{8.84}$&$75.79_{10.45}$&$76.81_{9.27}$&$69.63_{14.35}$
    \\
        
    GraphSAGE-EF
    &$73.21_{4.80}$&$76.06_{3.48}$&$80.80_{3.34}$&$74.21_{5,25}$&$78.43_{6.35}$
    \\
    GraphSAGE-LF
    &$66.82_{8.01}$&$70.78_{7.42}$&$72.41_{12.10}$&$70.34_{5.91}$&$73.69_{17.75}$
    \\

    GIN-EF
    &$67.16_{5.02}$&$75.44_{6.34}$&$76.74_{9.35}$&$76.86_{3.37}$&$72.40_{11.39}$
    \\
    GIN-LF
    &$68.43_{8.79}$&$74.28_{7.98}$&$72.35_{12.13}$&$78.27_{5.48}$&$66.67_{15.41}$
    \\
    
    BrainNetCNN-EF
    &$61.82_{12.12}$&$66.61_{5.71}$&$68.17_{8.40}$&$66.70_{3.33}$&$65.08_{10.91}$
    \\
    BrainNetCNN-LF
    &$73.35_{6.85}$&$69.56_{3.48}$&$70.96_{4.52}$&$72.07_{3.61}$&$62.47_{4.81}$
    \\   
    
    BrainGNN-EF
    &$64.22_{11.03}$&$64.39_{10.30}$&$63.44_{16.00}$&$67.82_{8.38}$&$62.30_{16.02}$
    \\ 
    BrainGNN-LF
    &$72.05_{5.09}$&$74.11_{6.67}$&$77.52_{7.22}$& $73.72_{6.96}$ & $73.77_{7.76}$ 
    \\ 
    
    HGCN-EF
    &$65.29_{9.27}$&$63.61_{4.20}$&$63.63_{8.03}$&$67.27_{3.14}$&$55.87_{8.29}$
    \\    
    HGCN-LF
    &$67.69_{4.38}$&$66.39_{4.12}$&$68.82_{6.56}$&$68.93_{2.90}$&$60.53_{9.72}$
    \\
    
    \cmidrule{1-6}
    HKGF$_1$~(Ours)&$75.96_{2.37}$&$76.78_{3.44}$&$78.70_{4.82}$&$80.50_{1.07}$&$74.66_{10.01}$ 
    \\
    HKGF$_2$~(Ours)& $\bm{76.85}_{3.78}$ & $\bm{79.22}_{4.27}$ & $\bm{81.12}_{5.32}$ & ${79.69}_{4.36}$ &${75.04}_{8.77}$ 
    \\
    \bottomrule
    \end{tabular}
    \label{tab_SMCfMRIASL}
\end{table}

\if false
\begin{table}[!tbp]
\setlength{\abovecaptionskip}{0pt}
\setlength{\belowcaptionskip}{0pt}
\setlength{\abovedisplayskip}{0pt}
\setlength{\belowdisplayskip}{0pt}
    \centering
        \caption{Results (\%) of the proposed methods and their variants without model pretraining in ANI vs. CN classification on HAND. 
}
\renewcommand{\arraystretch}{1}
\setlength\tabcolsep{0.2pt}
\begin{tabular}{l|ccccccc}
\toprule
Method & AUC & ACC & F1 & BAC& SEN & SPE & PRE \\
\midrule
HKGF$_1$w/oP
&  $63.62_{3.48}$ & $68.03_{1.91}$ & $68.42_{4.13}$& $68.18_{1.85}$ & $\bm{72.81}_{8.47}$ & $63.54_{6.49}$ & $68.51_{1.54}$ \\

 HKGF$_1$~(Ours) & $\bm{68.83}_{2.55}$ & $\bm{71.53}_{2.59}$ & $\bm{69.72}_{5.15}$ &$\bm{71.54}_{2.53}$ & ${71.30}_{8.73}$ & $\bm{71.78}_{4.18}$ & $\bm{73.62}_{1.72}$  \\
\midrule 
\textcolor{blue}{HKGF$_2$w/oP }& $66.42_{2.52}$& $65.70_{2.79}$ & $66.69_{2.76}$ &  $66.12_{2.76}$& $72.46_{3.54}$ & $59.79_{5.06}$  & $65.57_{2.15}$ \\ 

    HKGF$_2$ (Ours)& $\bm{69.74}_{1.77}$ & $\bm{71.86}_{2.39}$ & ${68.73}_{2.54}$ &$\bm{72.36}_{1.40}$ &${70.45}_{7.57}$ & $\bm{74.27}_{7.64}$ & $\bm{76.45}_{4.99}$ \\

        \bottomrule
    \end{tabular}
\label{tab_pretrain}
\end{table}
\fi

\subsection{Model Generalizability Evaluation}
The proposed HKGF is a general framework that can be employed to fuse multiple modalities. 
To further validate its generalization ability on other modalities, we conduct a new experiment to perform SMC vs. CN classification using paired resting-state fMRI and arterial spin labeling (ASL) MRI data of 29 SMC and 15 CN subjects from ADNI. 
The prepossessing of the fMRI data is introduced in Section~\ref{S4_preprocessing}. 
We preprocess the ASL data using the ASLtbx toolbox~\cite{wang2008empirical}. 
From each preprocessed ASL MRI, we obtain a mean cerebral blood flow (CBF) map comprising 116 ROIs per subject, as defined by the AAL atlas. 
For each ROI, we extract a 107-dimensional radiomics feature vector from the CBF image using PyRadiomics~\cite{van2017computational}. 
Based on these regional features, we construct an ASL-based FC graph by treating each ROI as a node and computing the Pearson correlation between feature vectors of ROI pairs to define the edge weights. 

For a fair comparison, all the competing deep learning methods and our HKGF share the same fMRI and ASL graph inputs, while three conventional machine learning methods (\ie, SVM, RF and XGBoost) employ each subject's concatenated fMRI-based FC network feature and ASL-based radiomics features as input. 
The results achieved by different methods in SMC vs. CN classification on ADNI with fMRI and ASL data are reported in Table~\ref{tab_SMCfMRIASL}. 
This table suggests that our HKGF$_1$ and HKGF$_2$ outperform other methods across most metrics. 
HKGF$_2$ achieves the highest AUC (76.85\%) and ACC (79.22\%), indicating robust performance in distinguishing SMC subjects and CNs. 
These results demonstrate that our HKGF framework generalizes effectively across diverse multimodal scenarios. 

\subsection{Computation Complexity Analysis}
\label{S5_5}

We further compare the computational complexity of the proposed HKGCN and HKGAT with the three most relevant methods (\ie, GCN, GAT, and HGCN). 
We evaluate the number of their trainable parameters and the total number of floating-point operations (FLOPs) in one forward pass when encoding each functional connectivity graph from resting-state fMRI and structural connectivity graph from DTI for ANI vs. CN classification on the HAND cohort.  
Using an NVIDIA RTX 4090 GPU, we report the results in Table~\ref{tab_cost}, which reveal several key  observations. 
\if false
- HGCN requires 31.69M and 17.82M FLOPs to encode DTI and fMRI features, respectively. Our HKGCN reduces them to 19.27M and 12.38M FLOPs, suggesting less computational complexity. 
- HGCN vs. GCN vs. HKGCN vs. HKGAT vs. GAT 
\fi 
These results demonstrate that our HKGCN is more efficient than the state-of-the-art HGAN, achieving significantly lower computational cost while retaining the same number of trainable parameters. 
Recalling the classification results in Tables~\ref{tab_SMCfMRIDTI}-\ref{tab_ANIfMRIDTI}, we can observe that our HKGCN consistently outperforms HGAN while requiring fewer resources, demonstrating its effectiveness and scalability. 
In addition, our HKGAT model maintains a similar computational cost in terms of FLOPs and the same number of trainable parameters as the standard GAT. Likewise, our HKGCN matches the conventional GCN in both FLOPs and parameter count.  
This demonstrates that our novel approach of introducing a hyperbolic kernel into GAT and GCN does not increase computational complexity, yet significantly boosts their performance.


\subsection{Limitations and Future Work}
Several limitations should be taken into consideration. 
\emph{On one hand}, although we visualize the learned features and highlight discriminative structural and functional connectivity patterns in the experiments, a potential limitation of the proposed HKGF framework is the reduced interpretability of its representations, due to the complexity of hyperbolic geometry. 
Future work could incorporate attention-based explanation mechanisms~\cite{vaswani2017attention} or visualization tools in hyperbolic space to further enhance model interpretability.  
\emph{On the other hand}, HKGF requires complete multimodal data (\eg, paired DTI and fMRI), and thus, cannot handle cases with missing modalities. 
Future work will include graph data imputation techniques~\cite{han2022g} to impute the missing data. 
\if false
\emph{Furthermore}, the limited number of target data in this work poses a challenge for effective training of deep learning models. While we currently utilize large-scale auxiliary data for model pretraining to alleviate this issue partly, we will further address this limitation by integrating advanced 
self-supervised learning approaches~\cite{liu2021self} to improve robustness. 
\emph{Additionally}, when applying the pre-trained model to handle small target data, we don't consider the distribution gap between source and target domains; 
Future solution: Applying established domain adaptation techniques such as adversarial learning or feature alignment to reduce the distribution difference between source and target data.
\fi

\begin{table}[!t]
 \setlength{\abovecaptionskip}{0pt} 
 \setlength{\belowcaptionskip}{0pt}  
 \setlength\abovedisplayskip{0pt}
 \setlength\belowdisplayskip{0pt}
 \renewcommand{\arraystretch}{0.6}
\centering
\caption{
Comparison of computational costs for five methods in encoding a single FC graph from resting-state fMRI and a single SC graph from DTI for ANI vs. CN classification on the HAND dataset. 
Param: parameter; 
FLOP: floating-point operation; 
MMac: million multiply-accumulate operations.} 
\scriptsize
\setlength{\tabcolsep}{3pt}
\begin{tabular}{@{}l|cc|cc@{}}
\toprule
\multirow{2}{*}{Method} & \multicolumn{2}{c|}{fMRI}                    & \multicolumn{2}{c}{DTI}                    \\ \cmidrule(l){2-5} 
                        & \multicolumn{1}{c|}{Param (K)}   & FLOPs (MMac) & \multicolumn{1}{c|}{Param (K)} & FLOPs (MMac) \\ \midrule
GCN~\cite{kipf2017gcn} 
& \multicolumn{1}{c|}{11.65}       & 6.16      & \multicolumn{1}{c|}{26.50}     & 9.61      \\
HGCN~\cite{chami2019hgcn}                    & \multicolumn{1}{c|}{11.65}       & 8.95 & \multicolumn{1}{c|}{26.50}     & 15.84      \\
HKGCN (Ours)            & \multicolumn{1}{c|}{11.65}
& 6.22      & \multicolumn{1}{c|}{26.50}
& 9.67      \\ \midrule
GAT~\cite{velivckovic2017graph}                     & \multicolumn{1}{c|}{46.72}       & 198.22     & \multicolumn{1}{c|}{106.11}    & 273.14    \\
HKGAT (Ours)            & \multicolumn{1}{c|}{46.72}       & 198.89     & \multicolumn{1}{c|}{106.11}    & 273.82    \\ \bottomrule
\end{tabular}
\label{tab_cost}
\end{table}
\section{Conclusion}
\label{S6}
This paper introduces HKGF, a novel framework for neurocognitive decline analysis using multimodal neuroimaging data. 
The HKGF embeds brain functional/structural connectivity graphs into hyperbolic space via a family of novel hyperbolic kernel graph neural networks, thereby capturing local and global dependencies while preserving the hierarchical structure of brain networks. 
A cross-modality coupling module further enhances the fusion of DTI and fMRI data. 
Extensive experimental results show that HKGF consistently outperforms state-of-the-art methods in predicting neurocognitive decline, highlighting its effectiveness and potential in objectively quantifying brain connectivity changes linked to neurocognitive decline.

\section*{Acknowledgments}
The authors would like to thank Ms. Biying Xiu for her assistance with ASL data processing in this project. 
A part of the data is from ADNI.   
ADNI investigators provide data but are not involved in data processing, analysis, and writing. 
A list of ADNI investigators is accessible \href{https://adni.loni.usc.edu/wp-content/uploads/how\_to\_apply/ADNI\_Acknowledgement\_List.pdf}{online}.

\footnotesize
\bibliography{references}
\bibliographystyle{IEEEtran}

\if false
\vspace{-36pt}
\begin{IEEEbiography}[{\includegraphics[width=1in,height=1.25in,clip]{Meimei.jpeg}}]{Meimei Yang} received her PhD degree in computer science and engineering at Southeast University, Nanjing, China, in 2024. She is currently a Postdoc fellow in the Department of Radiology and BRIC at the University of North Carolina at Chapel Hill (2024-present). 
Her current research primarily focuses on machine learning and neuroimaging analysis. 
\end{IEEEbiography}
\vspace{-36pt}
\begin{IEEEbiography}[{\includegraphics[width=1in,height=1.25in,clip]{HKGNN-TPAMI/figs/YonghengSun.jpeg}}]{Yongheng Sun} is currently a Research Assistant at the University of North Carolina at Chapel Hill (2025-present). His research focuses on pattern recognition and medical image analysis. 
\end{IEEEbiography}
\vspace{-36pt}
\begin{IEEEbiography}[{\includegraphics[width=1in,height=1.25in,clip]{HKGNN-TPAMI/figs/QianqianWang.jpeg}}]{Qianqian Wang} is currently pursuing her Ph.D. degree in Biomedical Engineering at the University of North Carolina at Chapel Hill (2025-present). Her research focuses on biomedical data analysis and machine learning. 
\end{IEEEbiography}
\vspace{-36pt}
\begin{IEEEbiography}[{\includegraphics[width=1in,height=1.25in,clip,keepaspectratio]{HKGNN-TPAMI/figs/Andrea Bozoki-new.jpeg}}]{Andrea Bozoki} is an experienced academic clinician. Her research focuses on clinical trials and multimodal imaging biomarker development and validation. She has demonstrated expertise in obtaining and completing sponsored clinical trials, and in creating a revenue-positive multidisciplinary program for care of cognitive disorders. 
\end{IEEEbiography}
\vspace{-36pt}
\begin{IEEEbiography}[{\includegraphics[width=1in,height=1.25in,clip,keepaspectratio]{HKGNN-TPAMI/figs/MaureenKohi.jpg}}]{Maureen Kohi} 
received her medical degree from New York Medical College and completed her Diagnostic Radiology residency at the University of California, San Francisco. 
She has extensive clinical experience in detecting imaging biomarkers and applying imaging techniques in clinical decision-making. 
\end{IEEEbiography}
\vspace{-36pt}
\begin{IEEEbiography}[{\includegraphics[width=1in,height=1.25in,clip,keepaspectratio]{HKGNN-TPAMI/figs/MX-July-06.jpeg}}]{Mingxia Liu}
received the Ph.D. degree from Nanjing University of Aeronautics and Astronautics, Nanjing, China, in 2015. She is a Senior Member of IEEE (SM'19). Her current research interests include machine learning, pattern recognition, and medical data analysis.
\end{IEEEbiography}
\fi

\end{document}